\documentclass[lettersize,journal]{IEEEtran}
\usepackage{amsmath,amsfonts}
\usepackage{algorithmic}
\usepackage{algorithm}
\usepackage{array}
\usepackage[caption=false,font=normalsize,labelfont=sf,textfont=sf]{subfig}
\usepackage{textcomp}
\usepackage{stfloats}
\usepackage{url}
\usepackage{verbatim}
\usepackage{graphicx}
\usepackage{cite}
\usepackage{booktabs} 
\usepackage{svg}
\usepackage{diagbox}
\usepackage{makecell} 

\begin{document}

\title{MLINE-VINS: Robust Monocular Visual-Inertial SLAM With Flow Manhattan and Line Features}

\author{
  Chao~Ye,
  Haoyuan~Li,
  Weiyang~Lin,
  Xianqiang~Yang
  \thanks{The authors are with Research Institute of Intelligent Control and Systems, Harbin Institute of Technology, Harbin 150001, China}
}



\maketitle

\begin{abstract}
  In this paper we introduce MLINE-VINS,  a novel monocular visual-inertial odometry (VIO) system that leverages line features and Manhattan Word assumption.
  Specifically, for line matching process, we propose a novel geometric line optical flow algorithm that efficiently tracks line features with varying lengths, whitch is do not require detections and descriptors in every frame.
  To address the instability of Manhattan estimation  from line features,  we propose a tracking-by-detection module that consistently tracks and optimizes Manhattan framse in consecutive images.
  By aligning the Manhattan World with the VIO world frame, the tracking could restart using the latest pose from back-end, simplifying the coordinate transformations within the system.
  Furthermore, we implement a mechanism to validate Manhattan frames  and a novel global structural constraints back-end optimization.
  Extensive experiments results on vairous datasets, including benchmark and self-collected datasets, show that the proposed approach outperforms existing methods  in terms of accuracy and long-range robustness.
  The source code of our method is available at: https://github.com/LiHaoy-ux/MLINE-VINS.
\end{abstract}

\section{Introduction}
\IEEEPARstart{A}{ccuracy} of pose estimation is a critical factor in various fields, such as autonomous driving, augmented reality, and robotics.
Simultaneous localization and mapping (SLAM) has proven to be an effective approach to address this challenge \cite{10304161,10042487}.
Among SLAM techniques, visual-inertial odometry (VIO) is  particularly popular  due to its cost-effectiveness, accuracy, and robustness.

In VIO, point feature is widely used for  camera pose estimation due to its simplicity and efficiency.
Representative point-based VIO systems include MSCKF-VIO\cite{sun2018robust}, OK-VINS\cite{leutenegger2016okvis}  and VINS-MONO\cite{8421746}, with VINS-MONO being one of the most  widely adopted algorithm.
However, the performance of point-based VIO is affected by the number and spatial distribution of points and it significantly hindered in textureless environments, where the lack of texture leads to point loss.
To address these limitations, line features are increasingly considered as a valuable complement to point features improving the robustness of VIO systems.

Line features arecommonly found in low-texture environments, particularly in man-made environments\cite{10616142}.
To enhance the robustness of VIO systems, researchers have incorporated line features into VIO systems, such as PL-VINS\cite{fu2020pl}, PL-VIO\cite{he2018pl}, UL-SLAM\cite{10488029}.
They demonstrate better performance compared to point-only systems in textureless environments.
However, a significant challenge in existing line-based VIO systems lies in the  efficiency of line-matching, which is often time-consuming and unsuitable for real-time applications.
Traditional methods rely on  LSD\cite{von2008lsd} to extract line features and LBD\cite{zhang2013efficient} to match line features in every frame, both of which are computationally expensive and impractical for real-time use.
Wang et al. \cite{wang2021line} proposed a deep learning-based line flow algorithm,  but it still requires to detect lines in every frame.
Xu et al. \cite{9998999}  introduced a geometric-based line tracking algorithm that achieves real-time performance. 
Compared to deep-learning based method, the method is more efficient do not require obtain detections and descriptors. 
However, it does not account for  changes in line segment lengths in across  frames.
These limitations restrict the effectiveness of current line optical flow algorithms.
To address these issues, we propose a novel line optical flow algorithm capable of tracking line features with varying lengths in consecutive frames.

Although line features enhance VIO performance in textureless environments, they fail to address the drift problem  during long-term operation.
The Manhattan world (\textit{MW}) assumption \cite{NIPS2000_90e13578}, a structural regularity in artificial environments, provides additional long-term information.
In such environments, planes or lines align with  three perpendicular predominant directions.
This characteristic has been widely used to decoup  rotation and translation  in visual  systems\cite{li2018monocular}.
However, few studies have leveraged \textit{MW} information to optimize VIO systems.
Some researchers\cite{zou2019structvio} indirectly incorporate  \textit{MW} constraints covertly by aligning line features with   the \textit{MW}'s axes, significant reductions in cumulative error.
Typically, the \textit{MW} is established using three orthogonal vanishing points  in camera frame\cite{7926628},  which are formed by the intersections of parallel lines in the image.
Researchers have explicitly utilized  vanishing points to construct  constraints for system optimization\cite{lim2022uv}.
However, extracting vanishing points is computationally expensive and  the global constraints between the \textit{MW} and line features are often overlooked.
In this paper, we propose a fast and robust  tracking-by-detection module to track Manhattan in consecutive frames.
For \textit{MW} validation, we introduce a pose-guided Manhattan verification method.
Additionally, we propose a novel back-end optimization framework that incorporates both local and global Manhattan constraints.

To address these issues, we propose a novel VIO system called MLINE-VINS integrates line features and the \textit{MW} assumption to enhance the robustness and accuracy of the system.
The primary contributions of this paper are as follows:
\begin{itemize}
  \item We propose a novel optical flow algorithm based on geometric to efficiently and reliably track line features with changes in length across consecutive frames, outperforming traditional methods.
  \item We propose an efficient tracking-by-detection module for tracking Manhattan in consecutive frames.  Initially, tracked lines are used to estimate the inital Manhattan, which is then refined with both tracked and supplementary lines. Additionally, a pose-guided Manhattan verification method is introduced for validation in the back-end.
  \item We estimate a novel global constraints back-end optimization framework that leverages  structural information among all frames.
  \item Extensive experiments  on  benchmark and collected datasets demonstrate that the proposed VIO system outperforms existing point-line monocular VIO systems.
\end{itemize}
\begin{figure}[!t]
  \centering
  \includegraphics[width=4in]{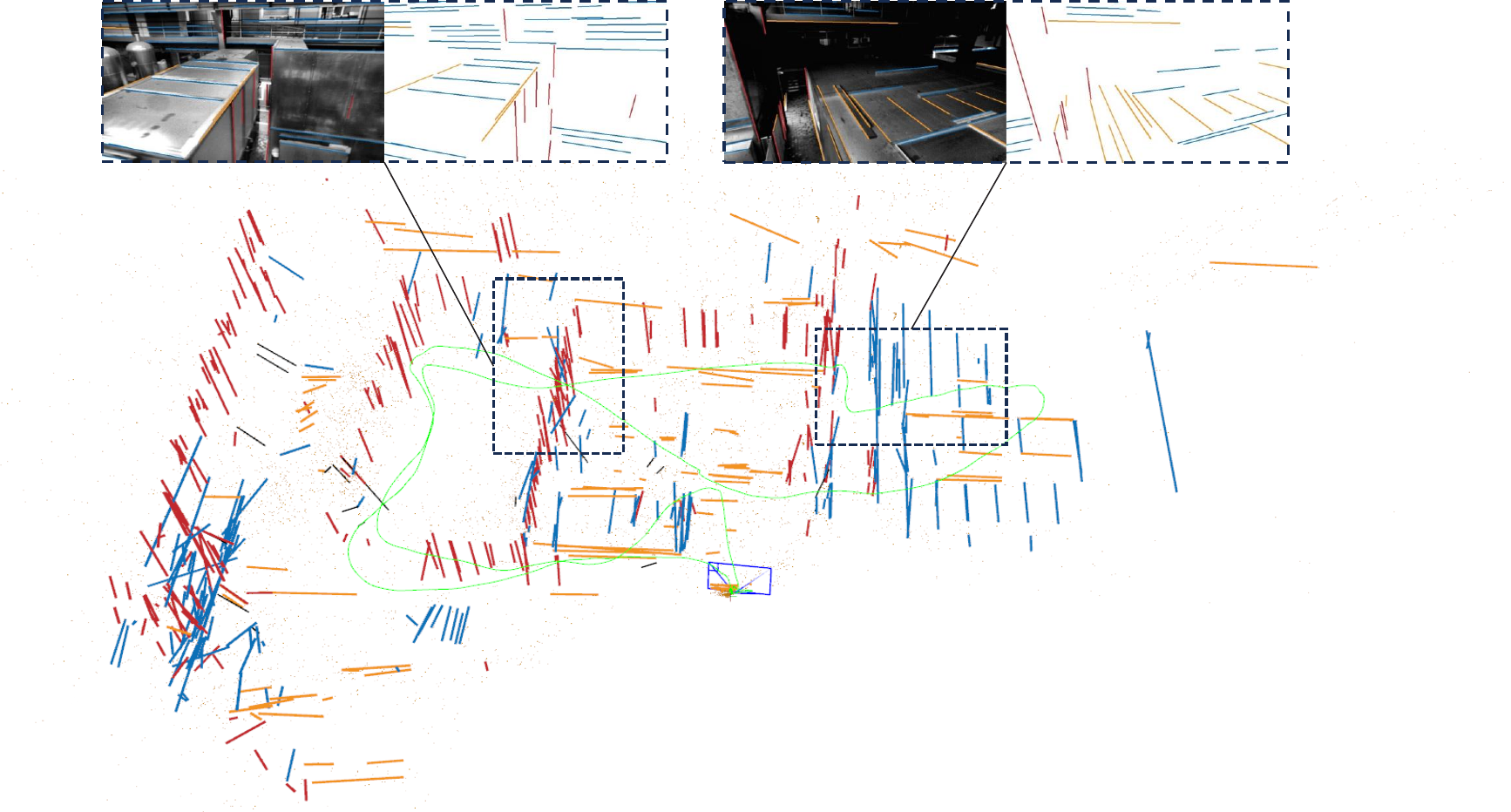}
  \caption{Results of the proposed VIO system on EuRoc \textit{MH05}.
    The blue, yellow, and red lines represent 3D structural lines along the X, Y, and Z directions, respectively.
    The green trajectory indicates the camera's historical poses.
    The black poins represent  3D points, and the 2D structural lines extracted from the RGB image are also shown at the top.}
  \label{fig8}
\end{figure}

\section{RELATED WORK}
In this work, we focus on pose estimation in structural environments uing monocular visual-inertial odometry.
Accordingly, we categorize  related studies into two areas: line based SLAM and  SLAM aided by structural regularity .
\subsection{Line Based SLAM}
Compared to points, line features are good components to enhance VIO performance in low-texture environments.
Therefore, many studies, sunch as PL-VIO\cite{he2018pl}, PL-VINS\cite{fu2020pl} have incorporated line features as landmarks in SLAM systems.
However, the efficiency of line matching remains a significant challenge for real-time applications in line-based SLAM systems.
Traditional methods rely on  geometric relationships \cite{schmid1997automatic} between  frames to identify candidate lines and use RANSAC\cite{chum2003locally} for selection.
These methods, however,  require accurate prior pose information, which conflicts with with the objectives of SLAM systems.
Alternatively, widely used descriptor-based matching methods like LBD\cite{zhang2013efficient} and MSLD\cite{wang2009msld} do not require prior information.
However, extracting descriptors is computationally expensive due to repetitive processing and cannot handle changes in line lengths.

Some researchers have applied the optical flow concept, originally used for point features,  to design line flow algorithms\cite{yan2023efficient}, often overlooking the geometric constraints of line features.
In\cite{wei2021point} and \cite{wang2021line}, the authors propose a method using line optical flow to guide line extraction from LSD in the next frame, which is time-consuming.
EPLF-VINS\cite{9998999} introduces a line flow tracking method based on the grayscale invariance hypothesis\cite{baker2004lucas}  and line geometric constraints, but it does not account for changes in the lengths of line segments.

\subsection{SLAM Aided by Structural Regularity }
The structural regularity of \textit{MW} is widely used to enhance SLAM system performance.
In VIO, \textit{MW} is  typically indirectly used to optimize the representation of line features, but it is not incorporated into the back-end optimization.
Struct VIO\cite{zou2019structvio} covertly incorporates MW constraints by aligning line features with the directions of the \textit{MW's} axes, achieving excellent performance..
But the MW constraints are only used in the front-end and are not utilized in the back-end optimization.
UV-SLAM\cite{lim2022uv} introduces additional  constraints between vanish points and linefeatures into the optimization problem.
This improves the performance of the VIO system.
However, these MW constraints are only applied within local optimization, neglecting the global constraints between \textit{MW} and line features.

Manhattan frames are typically estimated by clustering planar normal vectors.
For  purely visual scenarios, researchers use vanishing points  to indirectly get Manhattan.
The 2-line method\cite{7926628} clusters intersections of lines projected onto a sphere and selects the most reliable Manhattan.
While a Manhattan searching method\cite{li2020quasi} is also proposed to find Manhattan on a image.
However, these methods rely heavily on the number and distribution of line features, making them less robust in complex indoor environments.
Since these approaches estimate Manhattan from individual images, they fail to address the challenge of matching coordinate axes in consecutive frames.
Additionally, most Manhattan-based SLAM systems commonly use LSD for line extraction and the 2-line method for Manhattan estimation, which are both time-consuming and lack robustness.

In this paper, we take a step further to propose a novel VIO system by addressing the aforementioned issues.
Our line optical flow algorithm efficiently tracks line features across consecutive frames, accommodating changes in length.
Additionally, a novel tracking-by-detection module can continuous track Manhattan, which is fast and robust in complex indoor environments.
And a pose guided Manhattan frame verification is proposed to validate the Manhattan.
Finally, we leverage the \textit{MW} model to incorporate both global and local constraints in the back-end optimization.

\section{SYSTEM OVERVIEW }
\begin{figure*}[!h]
  \centering
  \includegraphics[width=0.99\textwidth]{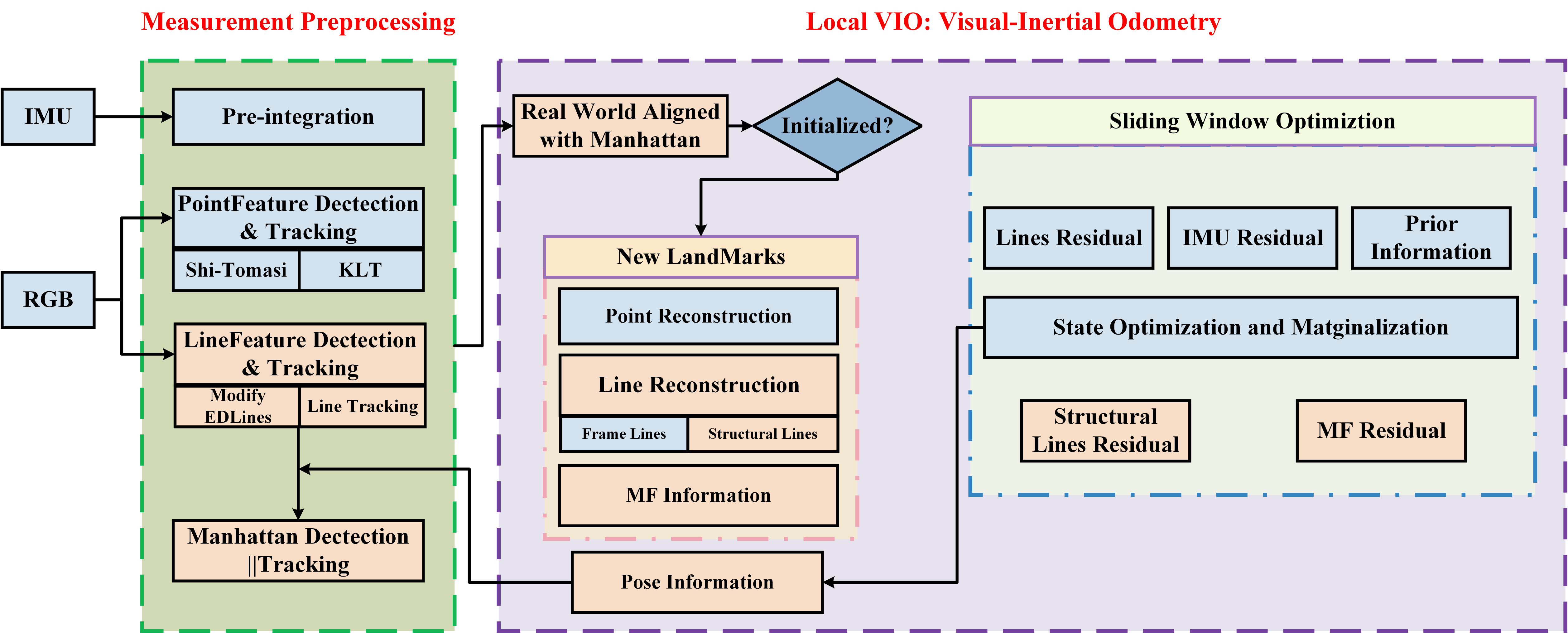}
  \caption{Overview of MLINE-VINS.
    The orange boxes highlight the new components introduced in this paper.
    Upon receiving the RGB input, points and lines are extracted and tracked in parallel.
    The Manhattan tracking-by-detection module is then executed to estimate \textit{MFs} in consecutive images, with each line being clustered according to a principal axis.
    If the \textit{MF} is lost, the last camera state from the back-end is used as the initial value to restart tracking.
    After system initialization, the  VIO world frame is aligned with the \textit{MW}, and optimization is performed  using the local and global constraints of Manhattan and structural lines.}
  \label{flow_chart}
\end{figure*}

This paper introduces MLINE-VINS, a novel monocular visual-inertial odometry system designed for robust performance in complex indoor environments.
The pipeline of our system is shown in Fig. \ref{flow_chart}.
We propose innovative strategies for Manhattan and line feature tracking, coupled with back-end optimization leveraging Manhattan constraints.
Our approach focuses on improving robustness and accuracy of the VIO system in in challenging indoor  environments.

\subsection{Notation and Manhattan Model}
In this paper, we use the IMU frame as the body frame, and the notations are defined in Table \ref{tab6}.
$\mathbf{R}\in SO(3)$, $\mathbf{t}\in \mathbb{R} ^3$ represent the rotation matrix and translation vector, respectively, and $ \mathbf{T}=\begin{bmatrix}
    \mathbf{R} & \mathbf{t} \\
    0          & 1
  \end{bmatrix}. $ denotes the transformation matrix.
\begin{table}[!h]
  \centering
  \caption{Some Important Notations}
  \label{tab6}
  \begin{tabular}{cc}
    \toprule
    \specialrule{0em}{3pt}{3pt}
    Notations    & Explanation                                         \\
    \specialrule{0em}{3pt}{3pt}
    \hline
    \specialrule{0em}{1pt}{1pt}
    $\times$     & Cross product operation                             \\
    \specialrule{0em}{1pt}{1pt}
    $(\cdot )^w$ & A vector $(\cdot )$ in global world frame           \\
    \specialrule{0em}{1pt}{1pt}
    $(\cdot )^c$ & vector $(\cdot )$ in camera  frame                  \\
    \specialrule{0em}{1pt}{1pt}
    $(\cdot )^b$ & vector $(\cdot )$ in IMU frame                      \\
    \specialrule{0em}{1pt}{1pt}
    $(\cdot )^M$ & vector $(\cdot )$ in Manhattan frame                \\
    \specialrule{0em}{1pt}{1pt}
    $T^c_b$      & The extrinsic matrix from IMU frame to camera frame \\
    \specialrule{0em}{1pt}{1pt}
    \bottomrule
  \end{tabular}
\end{table}

\begin{figure}[!h]
  \centering
  \includegraphics[width=2.5in]{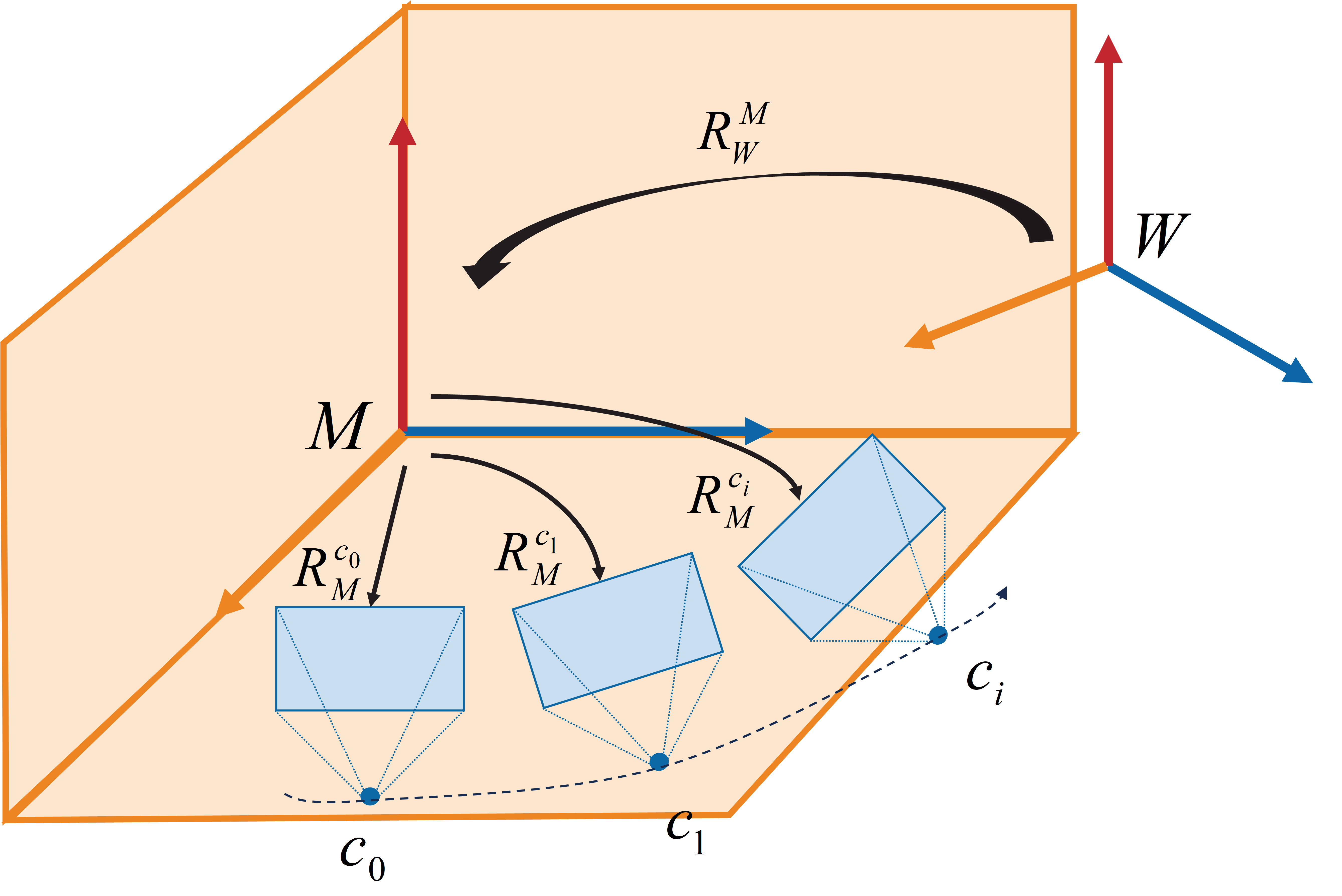}
  \caption{Rotation estimation between camera frame and \textit{MW}.
    The extrinsic matrix $R^M_W$ represents the rotation between VIO world frame and \textit{MW}.
    The rotation changes between \textit{MFs}  in camera coordinate  are represented by  $\mathbf{R}^{c_i}_M$. }
  \label{fig6}
\end{figure}

The \textit{MW} is a static box world with three orthogonal directions, known as the principal axes.
Most line features in the scene are aligned with one of these principal axes.
We define Manhattan frame observation in the camera frame as the Manhattan Frame (\textit{MF}) which can be expressed as $\mathbf{R}^{c_i}_M$.
As  shown in Fig. \ref{fig6}, the relative rotation between two \textit{MFs}, $\mathbf{R}^{c_j}_{c_i}$ can be easily computed as follows:
\begin{equation} \label{eq24}
  \begin{aligned}
    \mathbf{R}^{c_j}_{c_i}= \mathbf{R}^{c_j}_M  (\mathbf{R}^{c_i}_M )^{-1}
  \end{aligned}
\end{equation}
We apply the \textit{MW} assumption to enhance the system's robustness in complex indoor environments.

\subsection{Local Visual-Inertial Odometry}
The local VIO system consists of three components: measurement processing, system initialization, and nonlinear optimization.
\subsubsection{Measuremnt Processing}
The system receives RGB images and IMU messages as inputs. For images, point and line features are processed in parallel. When the first image is input, point and line features are extracted using  Shi-Tomasi\cite{323794} and modified EDLines algorithm.
If line features are available, the system  detects \textit{MFs} and classifies lines into three principal directions using the 2-line method \cite{7926628}.
In consecutive frames, point features are matched using the  KLT tracker\cite{suhr2009kanade} , line features are matched with an improved line optical flow method for , and \textit{MFs} are matched using a proposed tracking-by-detection module.
If  \textit{MW}  tracking fails, the last camera pose state in the back-end is used as the initial value to restart tracking
For IMU messages, preintegration is accumulated between keyframes, following the same approach as VINS-MONO\cite{8421746}.

\subsubsection{System Initialization}
After collecting sufficient information from the front-end,  the system  initializes the primary state, which include scaling parameter $s$, pose $\mathbf{T}_{b_i}$, gravity $\mathbf{g}$, velocity $\mathbf{v_i}$, line 3D representation $\mathbf{o_i}$ and IMU gyroscope bias $\delta \mathbf{b_w}$.

The transformation $\mathbf{T}^{c_0}_{c_i}$ is initialized  using the Structure from Motion (\textit{SFM}) method, leveraging visual features in a sliding window.
It is then converted  to $ \mathbf{T}^{c_0}_{b_i}$ using the extrinsic matrix $\mathbf{T}^c_b$.
By simultaneously considering SFM results and IMU preintegration, the gyroscope bias $\delta \mathbf{b_w}$ is estimated through rotation differences.
The gravity $\mathbf{g}$, velocity $\mathbf{v_i}$, and scale $s$ are determined by minimizing location and velocity residuals.
To simplify the relationship between \textit{MW} and real VIO world frame, the VIO world frame is aligned with the  \textit{MW}.
Consequently, the $ \mathbf{T}^{c_0}_{b_i}$ is transformed into $\mathbf{T}^{M}_{b_i}$.

For line features, a 3D line $\mathcal{L}=[\mathbf{n}^T, \mathbf{d}^T]^T  $ is parametered using Plücker representation.
This representation comprises the line direction $\mathbf{d}$ and normal vector $\mathbf{n}$ of the plane formed by the line and camera center.
It is initialized using Plücker matrix $L^{*}$, which is derived from the intersection of two planes $\mathbf{\pi}_1$, $\mathbf{\pi}_2$ as shown below:
\begin{equation} \label{eq18}
  \begin{aligned}
    \mathbf{L}^{*}= \begin{bmatrix}
      [\mathbf{d}]_{\times} & \mathbf{n} \\
      -\mathbf{n}^T         & 0
    \end{bmatrix} = \mathbf{\pi}_1 \mathbf{\pi}^{T}_2 - \mathbf{\pi}_2 \mathbf{\pi}^{T}_1
  \end{aligned}
\end{equation}
\subsubsection{Nonlinear Optimization}
Following the system initialization, the state is optimized within a sliding-window.

\textbf{Line Orthogonal Representation}: The Plücker representation facilitates spatial transformations of lines but is impractical  for optimization due to over-parameterization.
Therefore, a more concise representation is adopted for optimizing line features.
The orthogonal representation is defined as:
\begin{equation}\label{eq19}
  \begin{aligned}
    \mathbf{o}_i   = [\mathbf{\psi }_{i}, \phi _i]
  \end{aligned}
\end{equation}
where $\mathbf{\psi }_{i} = \begin{bmatrix}\frac{\mathbf{n}}{||\mathbf{n}||}&\frac{\mathbf{d}}{||\mathbf{d}||}  &\frac{\mathbf{n} \times \mathbf{d}}{||\mathbf{n} \times \mathbf{d}||} \end{bmatrix}$, and $\phi_i$ is the distance between the line and the camera center.

\textbf{Optimization State}
The full state vector $\mathbf{\chi}$ is defined as follows:
\begin{equation}\label{eq20}
  \begin{aligned}
    \mathbf{\chi}  = & [\mathbf{x}_0, \mathbf{x}_1, ..., \mathbf{x}_n, \lambda_0, \lambda_1, ..., \lambda_m, \mathbf{o}_0, \mathbf{o}_1, ..., \mathbf{o}_k] \\
    \mathbf{x}_i   = & [\mathbf{q}^{M}_{b_i}, \mathbf{t}^{M}_{b_i}, \mathbf{v}^{M}_{i}, \mathbf{b}_{a}, \mathbf{b}_{w}]                                     \\
    \mathbf{o}_i   = & [\mathbf{\psi }_{i}, \phi _i]
  \end{aligned}
\end{equation}
$\mathbf{q}^{M}_{b_i}$ is  the quaternion form of $R^M_{b_i}$.
Note that all states are estimated in the \textit{MW} frame.

\begin{figure}[!h]
  \centering
  \includegraphics[width=0.45\textwidth]{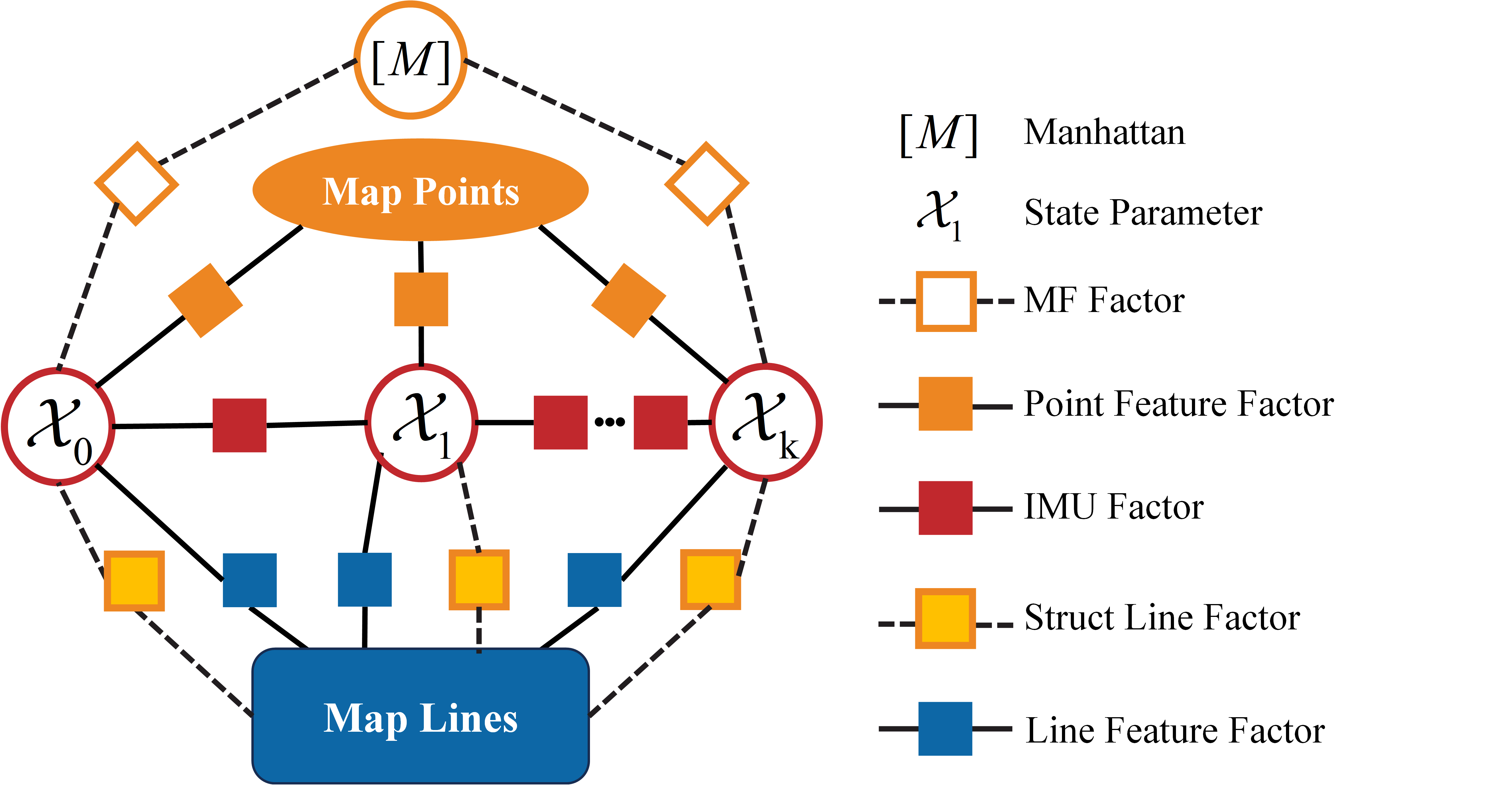}
  \caption{Factor graph for MLINE-VINS.
    The 'Map points' and 'Map lines'  nodes represent all point and line features within the sliding window.
    'Map Lines' include both structural and non-structural lines. Structural lines provide both reprojection and structural constraints, while non-structural lines only contribute reprojection constraints.}
  \label{factor_graph}
\end{figure}

The optimization problem is formulated as a visual-inertial adjustment and can be expressed as below:

\begin{equation}\label{eq21}
  \begin{aligned}
     & \min_{\chi}  \{||r_p - \mathbf{H}_p \chi||^{2}
    + \sum_{i \in \mathcal{B} }||r_{ \mathcal{B}}(z^i_{i+1}, \chi)||^2_{Cov^{b_i}_{b_{i+1}}}                                  \\
     & +  \sum_{(i, j) \in \mathcal{P}}\rho_p||r_{\mathcal{P}}(z^{c_i}_{P_j}, \chi)||^2_{Cov^{c_i}_{P_j}}
    + \sum_{(i, k) \in \mathcal{L}}\rho_l||r_{\mathcal{L}}(z^{c_i}_{L_k}, \chi)||^2_{Cov^{c_i}_{L_k}}                         \\
     & + \sum_{(i, m) \in \mathcal{M}}||r_{\mathcal{M}}(z^{c_i}_{M_m}, \chi)||^2_{Cov^{c_i}_{M_m}}                            \\
     & + \sum_{(i, n) \in \mathcal{L}_{struct}}||r_{\mathcal{L}_{struct}}(z^{c_i}_{L_m}, \chi)||^2_{Cov^{c_i}_{L^{struct}_m}}
    \}
  \end{aligned}
\end{equation}
the Huber norm $\rho_p$ \cite{huber1992robust} is defined as:
\begin{equation}\label{eq22}
  \begin{aligned}
    \rho_p(x) = \begin{cases}
      x            & x\le 1 \\
      2\sqrt{x} -1 & x> 1
    \end{cases}
  \end{aligned}
\end{equation}
where $r_{\mathcal{B}}$, $r_{\mathcal{P}}$, $r_{\mathcal{L}}$, $r_{\mathcal{M}}$, $r_{\mathcal{L}_{struct}}$ are residuals for IMU, point, line, Manhattan and structural line features respectively.
$\mathcal{P}$, $\mathcal{L}$ are 3D points and lines triangularized in the sliding window,
while  $\mathcal{M}$ denotes the \textit{MF} observation and $\mathcal{L}_{struct}$  refers to the structural lines among the set of lines $\mathcal{L}$.
The covariance matrix of the measurement is denoted as $Cov$, and $z$ represents the measurement.
The prior message from the marginalization of old states is represented as $[r_p,\mathbf{H}_p]$.
The factor graph for the defined cost function is shown in Fig. \ref{factor_graph}.
If there are no \textit{MFs} or structural lines, only point and line residuals are used for optimization.
The Ceres Solver\cite{agarwal2012ceres} is employed to solve the nonlinear optimization problem.

\section{IMPLEMENTATION OF KEY COMPONENTS}
\subsection{Novel Line Optical Flow  Algorithm}
Previous methods for line features dectection and matching are often time comsuming and inefficient.
Common approache, such as using LSD for line detection and LBD for descriptor-based matching,  are time-consuming and affected by repetitive textures.
To address these issues, this paperintroduces a novel line optical flow algorithm.
The algorithm operates in two stages: 1) Line Feature Initialization; 2) Line Optical Flow Tracking.
By eliminating the need to detect and match lines in every frame, the proposed approach is highly suitable for real-time applications.

\subsubsection[short]{Line Feature Initialization} In the first frame, a modified EDLines (\textit{MED}) is used to extract line features.
For robust feature tracking, line features are uniformly distributed across the image and co-linear features are merged.
If the number of line features is below the set threshold, the  grayscale threshold parameter of \textit{MED} is reduced, and line extraction is repeated..

\subsubsection[short]{Line Optical Flow Tracking} The primary concept is based on the grayscale invariance assumption. At time $ t+\delta t $,  a pixel $(u,v)$   at time $t$
moves to $(u',v')$, while maintaining a constant grayscale value, expressed as:
\begin{equation} \label{eq1}
  I(u, v, t) = I(u+\delta u, v+\delta v, t+\delta t)
\end{equation}

Using the Taylor expansion on the  right-hand of Eq. \eqref{eq1}, we obtain:
\begin{equation} \label{eq2}
  I(u, v, t)  = I(u, v, t) + \frac{\partial I}{\partial u} \delta u + \frac{\partial I}{\partial v} \delta v + \frac{\partial I}{\partial t} \delta t
\end{equation}
simplify the above equation, we obtain:
\begin{equation} \label{eq3}
  \bf{I_t} = - \left [ \begin{matrix} I_u   & I_v \end{matrix} \right ] \left [ \begin{matrix} \frac{du}{dt} \\ \frac{dv}{dt} \end{matrix} \right ]
\end{equation}
where $\bf{I_t}$ represents the gary temporal change in image, $I_u$ and $I_v$ denote the spatial intensity gradients at the  point.
Evey pixel in frame $F_t$ adheres to this relationship.
An additional constraint for line features is introduced below.

A line feature $l_i$ at time $t$ is defined as:
\begin{equation}
  \mathbf{l_i} = [(u_s, v_s), (u_1, v_1),...,(u_i, v_i), (u_e, v_e)]
\end{equation}
where $(u_s, v_s)$ is start point, and $(u_e, v_e)$ is endpoint.

\begin{figure}[!t]
  \centering
  \includegraphics[width=0.4\textwidth]{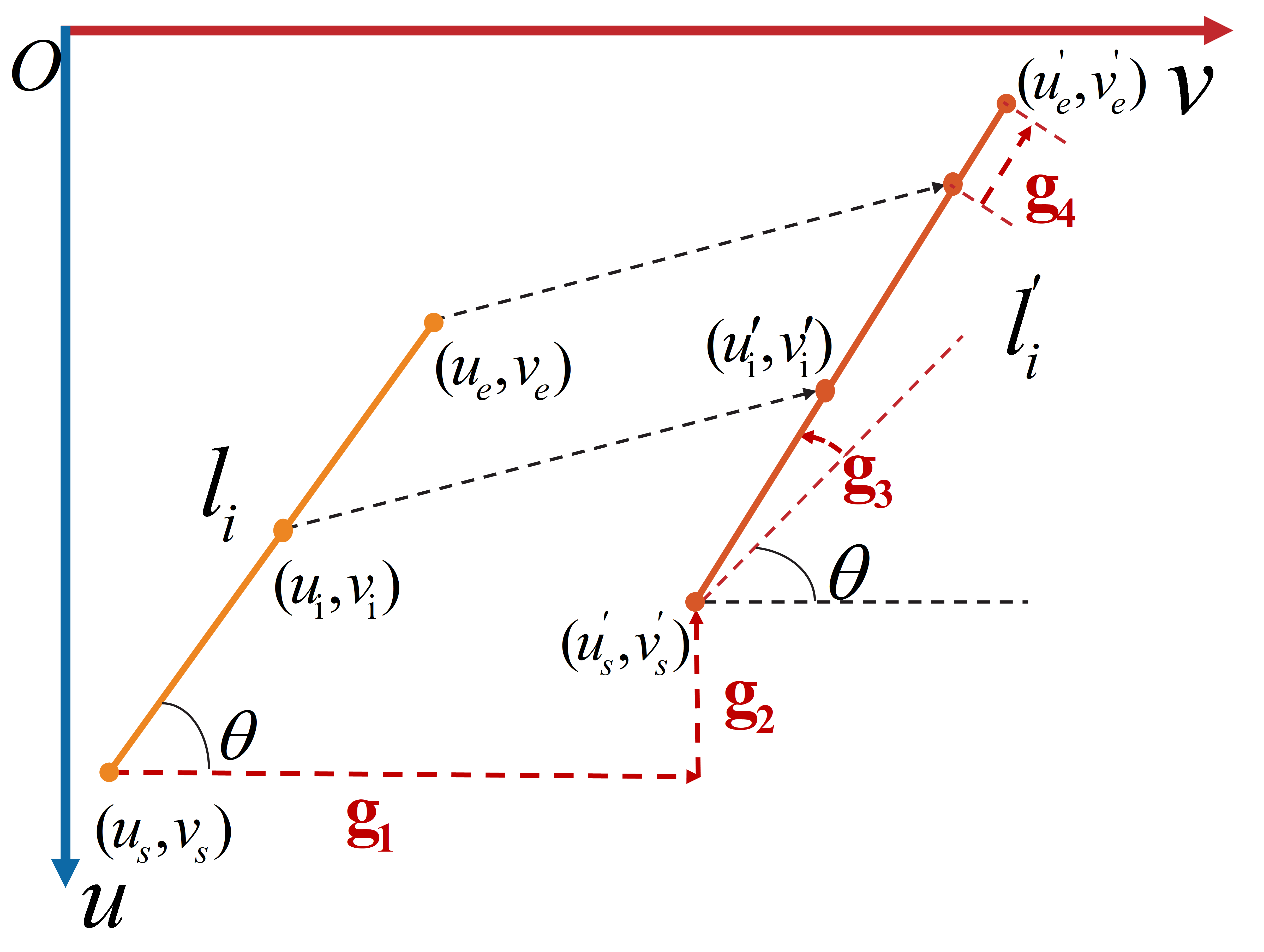}
  \caption{Diagram of Line feature tracking model.
    $l_i$ represents a line feature in frame $F_t$, and $l_i'$ is the tracked line feature in frame $F_{t+\delta t}$.
    $g1$, $g2$, $g3$, $g4$ denote the changes in the horizontal coordinate, vertical coordinate, angle and length changes of the line feature, respectively.}
  \label{fig1}
\end{figure}

As shown in Fig. \ref{fig1}, during $\delta t$, $l_i$ moves to $l_i'$ on the image.
The relationship between the same line in two consecutive frames is expressed as follows:
\begin{equation} \label{eq4}
  \begin{aligned}
    u_i' = u_i + g_1 + (l_i+g_4 l_i)cos(\alpha + g_3) -l_n cos \alpha
    \\
    v_i' = v_i +g_2 + (l_i+g_4 l_i) sin(\alpha + g_3) - l_n sin\alpha
  \end{aligned}
\end{equation}
where $g_1$ and $ g_2$ represent the position change of the starting point, $g_3$ denotes the angle change, and $g_4$ indicates the  change in line length.

By simplifying the equation and neglecting second-order small quantities, the equation can be rewritten as follows:
\begin{equation} \label{eq5}
  \begin{aligned}
    u_i' = u_i + g_1 + g_3(v_i - v_s) +g4(u_i - u_s) \\
    v_i' = v_i + g_2 - g_3(u_i - u_s) + g_4(v_i - v_s)
  \end{aligned}
\end{equation}

From Eq. \eqref{eq5}, we can derive:
\begin{equation} \label{eq6}
  \begin{aligned}
    \frac{du_i}{dt} = \frac{dg_1}{dt} - \frac{dg_3}{dt}(v_i - v_s) +  \frac{dg_4}{dt}(u_i - u_s)
    \\
    \frac{dv_i}{dt} = \frac{dg_2}{dt} - \frac{dg_3}{dt}(u_i - u_s) +  \frac{dg_4}{dt}(v_i - v_s)
  \end{aligned}
\end{equation}

Then we define $\frac{dg_1}{dt}$, $\frac{dg_2}{dt}$, $\frac{dg_3}{dt}$ and $\frac{dg_4}{dt}$,  as $\delta g_1$, $\delta g_2$, $\delta g_3$ and $\delta g_4$,respectively.
Consequently, Eq.\eqref{eq6} can be rewritten as:
\begin{equation} \label{eq7}
  \left [ \begin{matrix} u & v\end{matrix} \right ]
  =
  \left [\begin{matrix}
      1 & 0 & -(v_i - v_s) & u_i - u_s \\
      0 & 1 & (u_i - u_s)  & v_i - v_s
    \end{matrix} \right ]
  \left [\begin{matrix}
      \delta g_1 \\ \delta g_2 \\  \delta g_3 \\  \delta g_4
    \end{matrix} \right ]
\end{equation}

Each point on the line $l_i$ satisfies  Eq.\eqref{eq7}, forming an overdetermined linear equation with  $\delta g_1$, $\delta g_2$, $\delta g_3$ and $\delta g_4$.
This problem can be solved by using the Gauss-Newton method.
To enhance  feature tracking accuracy, the algorithm is integrated into a pyramid multi-scale optical flow framework.
Longer lines are considered  more reliable for feature tracking, so co-linear short lines are merged.
Based on the gradient, the starting  and endpoints of the lines are extended or merged.
If the number of tracked lines is insufficient, \textit{MED} is executed to extract additional features.

\subsection{Tracking-By-Detection Algorithm for Manhattan}
The runtime and accuracy of \textit{MFs} detection significantly influence the system's overall performance.
The number and distribution of structural lines primarily influence the effectiveness of Manhattan detection.
To address this, we propose an algorithm that incorporates line-tracking results to track \textit{MFs} in consecutive frames.
This method is designed to  improve the system's robustness in complex indoor environments.

\subsubsection{Initialization}
The detection module involves two essential steps: Dectecting  \textit{MF} during the inital stage, and selecting the most reliable one.
The 2-lines\cite{7926628}  method is employed to identify all potential  \textit{MFs} in a given frame, and then  the most reliable one is selected.
Since the result depends on the number and distribution of line features, validity is verified using the z-axis of the \textit{MF} and vertical lines.
If all conditions are satisfied,, the \textit{MF} $R^{c_i}_M$ is initialized.
\subsubsection{Tracking}
\begin{figure}[!t]
  \centering
  \includegraphics[width=0.48\textwidth]{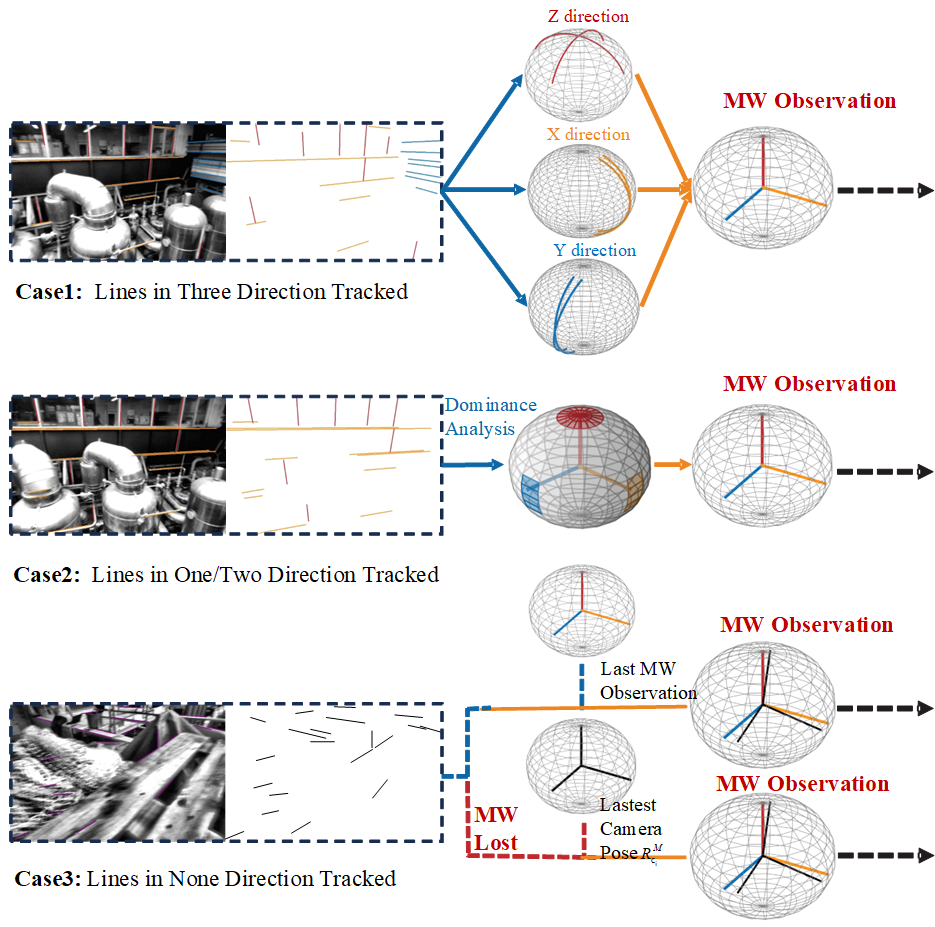}
  \caption{Illustration of Manhattan Tracking module.
    Case 1: Structural lines tracked in three directions. \textit{MF} is estimated by three vanishing points, formed by the intersection of lines in three directions  on a normal sphere.
    Case 2: Structural lines tracked in one or two directions.  \textit{MF} is estimated by the polar grid with weights related to length of lines.
    Case 3: Structural lines tracked failure. If the Manhattan exists in the previous frame, optimize it with current lines;  Otherwise, optimize \textit{MF} using the last camera position from back-end with lines.}
  \label{Manhattan_tracking}
\end{figure}

Since the original 2-lines method is not robust in scenarios with insufficient line features, we propose a tracking-by-detection module to track \textit{MF} in consecutive frames.
As  illustrated in Fig. \ref{Manhattan_tracking}, this module utilizes  prior  structural lines information from the previous frame to estimate \textit{MF} in the current frame.
The three principal axis directions of the \textit{MF} can be interpreted as the spatial positions of vanishing points.
\begin{equation}\label{eq12}
  \mathbf{R}^{c_i}_M =  \begin{matrix} [vp^{c_i}_1 & vp^{c_i}_2 & vp^{c_i}_3] \end{matrix}
\end{equation}

An illustration of the proposed algorithm is presented in Fig. \ref{Manhattan_tracking}.
This module classifies tracking scenarios into three categories based on the number of structural lines.

\textbf{Structural lines tracked in three directions:}
If the number of lines tracked in three principal directions is $n_1$, $n_2$ and $n_3$ ($n_i >2, i=1,2,3$),
the \textit{MF} in  the current frame can directly estimate an initial value through  the intersection of any two lines on a unit sphere.
The initial value  is defined as ${R^{c_i}_M}' = [vp_1', vp_2', vp_3']$.
Subsequently, structural line features are used to optimize this value.

The start  and end points of a line $l_i$ in the image are defined as $[u_{i_s}, v_{i_s}]$ and $[u_{i_e}, v_{i_e}]$, respectively.
The representation of a 3D line can be obtained using the equation below:
\begin{equation}\label{eq13}
  \begin{aligned}
    p^{c_i}_{i_s} & = \begin{matrix} [u_{i_s} & v_{i_s} & 1] \end{matrix}     \\
    p^{c_i}_{i_e} & = \begin{matrix} [u_{i_e} & v_{i_e} & 1]\end{matrix}     \\
    l^{c_i}_{i}   & = p^{c_i}_{i_s} \times p^{c_i}_{i_e}
  \end{aligned}
\end{equation}

The relationship between $l_i$ and its corresponding vanishing point direction is parallel.
Therefore,  ${R^{c_i}_M}'$ can be optimized using:
\begin{equation}\label{eq14}
  \begin{aligned}
    error = \sum_{i=1}^{n_1} l^{c_i}_{i} * vp_1 + \sum_{i=1}^{n_2} l^{c_i}_{i} * vp_2 + \sum_{i=1}^{n_3} l^{c_i}_{i} * vp_3
  \end{aligned}
\end{equation}

\textbf{Structural lines tracked in one or two directions:}
If lines are traced in only one or two directions, the scoring strategy for potential vanishing points is modified.
In 2-lines method, intersections of every two lines are mapped to a polar grid $\mathcal{G}$ with weights proportional to length.
And the best\textit{MF} is choosen  based on the highest score.
Structural lines are more reliable than common lines for identifying the \textit{MFs}.
For  a point $p$ on an unit sphere, its latitude  $\phi_1$ and longitude $\phi_2$ can be calculated, where $\phi_1 \in [0, \pi/2]$  and $\phi_2 \in [0, 2\pi]$.
The polar grid $\mathcal{G}(i, j)$ is initialized with a size of $90 \times 360$, and  the corressponding grid $\mathcal{G}(i, j)$ is updated using the following equation:
\begin{equation}\label{eq15}
  \mathcal{G}(i, j)  = \mathcal{G}(i, j) + ||\lambda l_1||*||\lambda l2||*sin(2 \theta)
\end{equation}
where $\lambda$ is an adjustable weight, if the lines $l_1$, $l_2$ are structural lines, $\lambda = 10$; otherwise, $\lambda = 1$.

\textbf{Structural lines tracked failure}
When no structural lines are  successfully tracked, the \textit{MF} from the  previous frame is used as the initial value to categorize lines in $n_1$, $n_2$ and $n_3$.
This approach is based on the assumption that the \textit{MF} changes minimally between two adjacent frames.
If the \textit{MF} is unavailable in the previous frame, the lastest camera pose  from the back-end is used as the initial value.
This is feasible because the \textit{MW} is aligned with the VIO world frame, as explained in the next section.
The \textit{MF} is then optimized according to Eq.\eqref{eq14}

The updated $vp_1$, $vp_2$, $vp_3$ are used to compose $R^{c_i}_M$, ensuring that  $R^{c_i}_M$ retains the properties of a valid rotation matrix.
Since the 2-lines method is inherently random and disorder, the order and reliability of the \textit{MF} need to be verified.
By incorporating these approaches, the tracking-by-detection module achieves consistency and reliability even in complex indoor environments.

\subsection{VIO World Frame Algin with Manhattan World}

\begin{figure}[!t]
  \centering
  \includegraphics[width=3.5in]{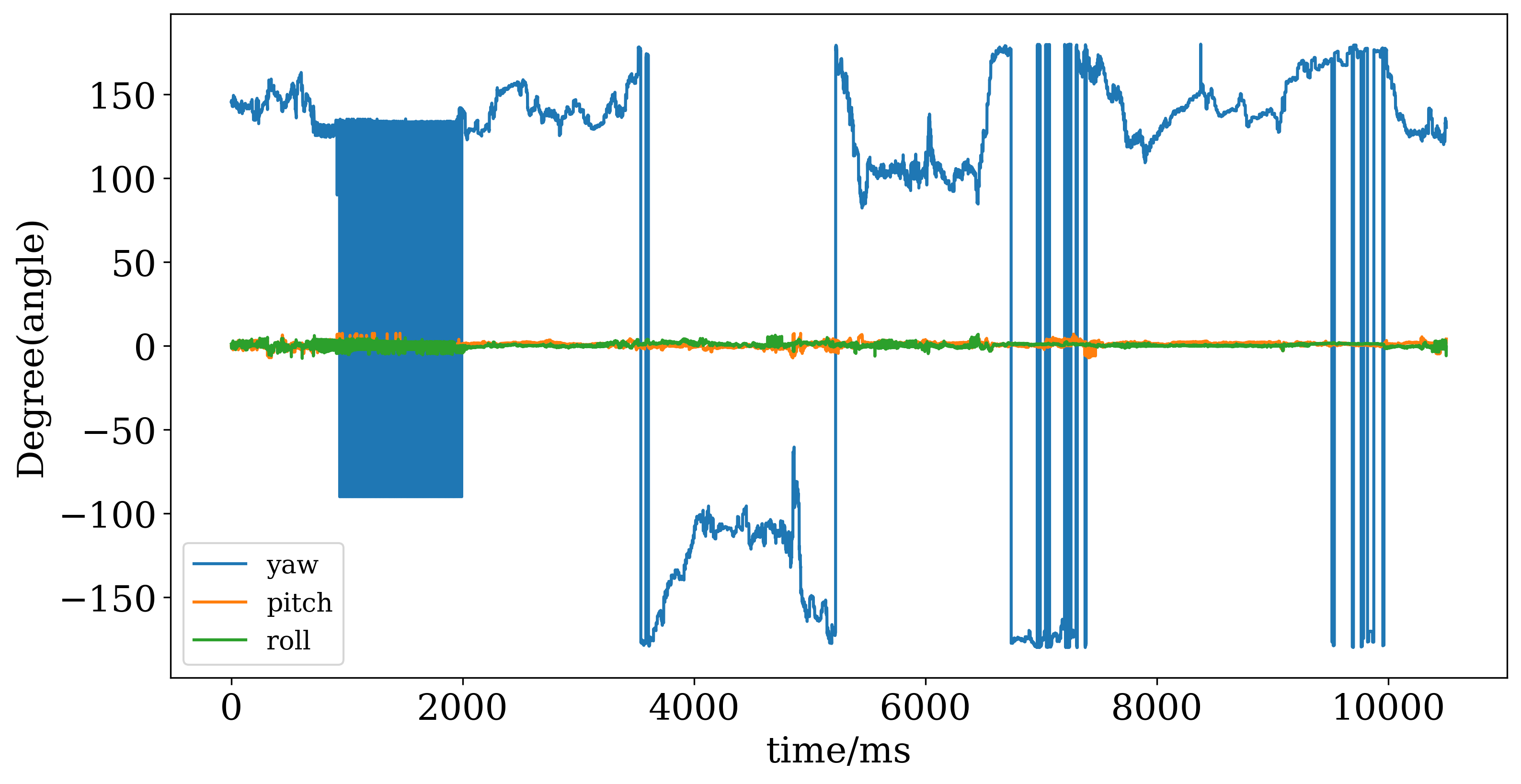}
  \caption{The angle error between VIO world frame and \textit{MW}.
  }
  \label{fig2}
\end{figure}

Since the \textit{MW} and VIO world frames are not aligned in the same coordinate system, we investigate the relationship between them.
In a sequence such as \textit{MH01},  ten consecutive frames are used for initialization to get VIO world frame.
To obtain the \textit{MW}, the \textit{MF} in the camera frame is transformed into the IMU frame.
The angular error between the VIO world frame and \textit{MW} is illustrated in Fig.\ref{fig2}.
Due to the yaw being unobservable during VINS-MONO initialization, its value is unstable.
The differences in roll and pitch are minimal, indicating that the z-axis direction of the \textit{MW} and VIO world frame are parrallel.
Furthermore, there is an implicit characteristic in indoor environment where building plumb lines are parallel to the direction of gravity.
Consequently, the z-axis of the \textit{MW} corresponds to the direction of gravity.
After VIO initialization, the body frame is expressed  in the world frame as $R^W_{b_i}$.
In the camera frame, the \textit{MW} is expressed as $R^M_{c_i}$.
The relationship between these two coordinate systems is given as follows:
\begin{equation}\label{eq8}
  \mathbf{R}^M_W =  \mathbf{R}^M_{c_i}  \mathbf{R}^c_b ( \mathbf{R}^W_{b_i})^{-1}
\end{equation}

Thus, the poses can be transformed from the VIO world frame to the \textit{MW} using the following relationship:
\begin{equation}\label{eq9}
  \mathbf{R}^M_{b_i} =  \mathbf{R}^M_W  \mathbf{R}^W_{b_i}
\end{equation}

After the VIO world frame is aligned with the \textit{MW}, it becomes easier to utilize the characteristics of the \textit{MW} and restart tracking using the state from the back-end.
Additionally, the yaw angle can also be observed through the \textit{MW}.
\subsection{Pose Guided Manhattan Frame Estimation Verification}
However, the results of MF estimation are not always reliable, and incorrect results can negatively impact rotation estimation.
To enhance system robustness, verification of the \textit{MFs} is necessary.
The pose error of VIO remains small over a short period within the sliding window.
Considering two frames $F_i$ and  $F_k$ in the VIO local window, the relative rotation between them should approximately equal the ground truth, $\mathbf{R}^{MF}_{i,k} \approx  \mathbf{R}^{VIO}_{i,k}$.
The reliability of the   \textit{MFs} is assessed using this relationship.

The local window consists of consecutive frames, i.e., $F_{k-n}$, $\dots$, $F_{k+n}$.
The  relative rotation between $F_i$ and other frames in window, $ \Delta  \mathbf{R}^{MF}_{i,k}$ and $\Delta  \mathbf{R}^{VIO}_{i,k}$ are computed using \textit{MFs} and  VIO, respectively.

Using Eq.\eqref{eq10}, the average angular error can be calculated as follows:
\begin{equation}\label{eq10}
  error_{ \mathbf{R}} = \frac{1}{2n} \sum_{i = k-n}^{k+n} \textbf{\textit{Angle}}( \Delta  \mathbf{R}^{MF}_{i,k} (\Delta  \mathbf{R}^{VIO}_{i,k})^{-1})
\end{equation}
where the function of \textbf{\textit{Angle}} is defined as follows:
\begin{equation}\label{eq11}
  \textbf{\textit{Angle}}_{ \mathbf{R}} = arcos(min(1, max(-1, \frac{trace( \mathbf{R}) -1}{2})))
\end{equation}

If the $error_{ \mathbf{R}}$ is below than the threshold $D_{Angle}$(e.g., $D_{Angle} = 0.5$ in this paper), the \textit{MF} is deemed reliable. Otherwise, it is considered unreliable and
the \textit{MF} will be discarded.
\subsection{Line Measuremnt Model}
For a 3D line $\mathcal{L}_k$ , we first transform it's Plücker representation from the Manhattan frame $(\cdot )^M$ to the camera frame $(\cdot )^{c_i}$ as follows:
\begin{equation}
  L^{c_i}_k = \begin{bmatrix} \mathbf{n_c} \\ \mathbf{d_c}\end{bmatrix} = T^{c_i}_M * \mathcal{L}_k=\begin{bmatrix}
    \mathbf{R}^{c_i}_M & [t^{c_i}_M ]_{\times} \mathbf{R}^{c_i}_M \\
    0                  & \mathbf{R}^{c_i}_M
  \end{bmatrix}\begin{bmatrix} \mathbf{n_w} \\ \mathbf{d_w}\end{bmatrix}
\end{equation}
Then, we use the following equation to project the line segment onto the image plane:
\begin{equation}
  \mathbf{l}_k = \begin{bmatrix} l_1 \\ l_2 \\l_3 \end{bmatrix}=\mathbf{K}_{proj}\mathbf{n}^{c_i}_k
\end{equation}
where $\mathbf{K}_{proj}$ represents the  projection matrix and $\mathbf{n}^{c_i}_k$ denotes  the direction of normalized line in $\{ c_i\}$.
As the image is in the normalized coordinate,  $\mathbf{K}_{proj}$ is an identity matrix.

\begin{figure}[!t]
  \centering
  \includegraphics[width=0.4\textwidth]{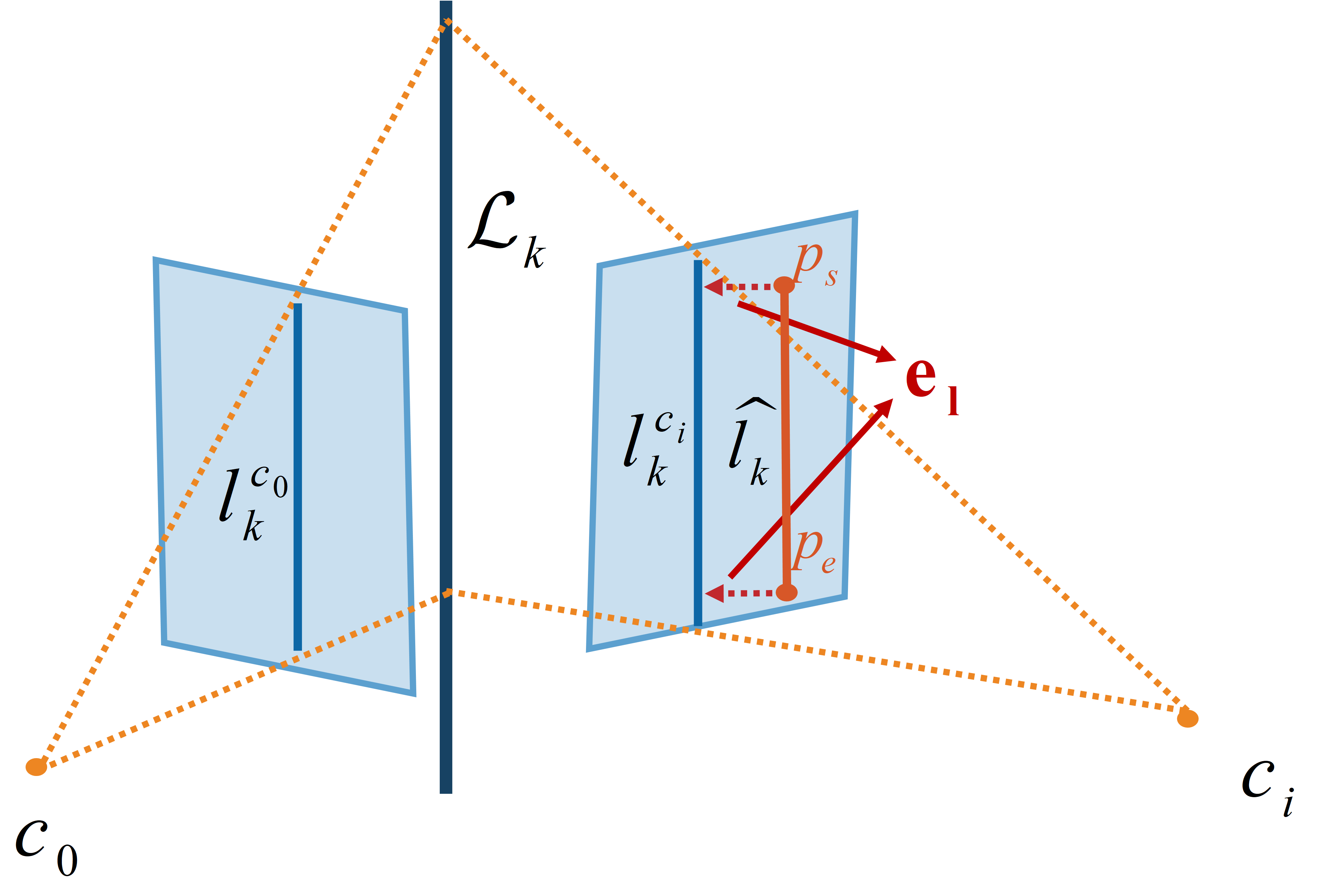}
  \caption{Representation of line residual model.
    The blue line represents the reprojection of the 3D line in the image, while the orange line represents the observation.
    The residual $\mathbf{e_l}$ is calculated as the distance between the two lines.}
  \label{line_residual}
\end{figure}
As illustrated in Fig.\ref{line_residual}, the line residual is defined as the distance between the projected line and the observed line on the image plane.
This relationship can be expressed as:

\begin{equation}\label{eq23}
  \mathbf{e}_l = \begin{bmatrix}
    d(\mathbf{l}_s, \mathbf{l}_k) \\d(\mathbf{l}_e, \mathbf{l}_k)
  \end{bmatrix}
\end{equation}
where
\begin{equation}
  d(\mathbf{l}_s, \mathbf{l}_k) = \frac{|\mathbf{l}_s^T \mathbf{l}_k|}{||\mathbf{l}_k||}
\end{equation}
$\mathbf{l}_s = (u_s, v_s, 1)$ and $\mathbf{l}_e = (u_e, v_e, 1)$ represent the start and endpoint of observed line,respectively and let $d$ denote the  distance between point and re-projected line.
The corressponding Jacobian $J_l$ can be obtained  using the chain rule:
\begin{equation}
  J_l = \frac{\partial e_l}{\partial n^{c_i}_k}\frac{\partial n^{c_i}_k}{\partial  L^{c_i}_k}
  \begin{bmatrix}
    \frac{\partial L^{c_i}_k}{\partial \delta x_i} & \frac{\partial L^{c_i}_k}{\partial \mathcal{L}_k} \frac{\partial \mathcal{L}_k}{\partial \delta o_i}
  \end{bmatrix}
\end{equation}
with
\begin{equation}
  \refstepcounter{equation}
  \begin{aligned}
    \frac{\partial e_l}{\partial n^{c_i}_k}
     & =  \begin{bmatrix}\frac{\partial e_l}{\partial n^{c_i}_k} & \frac{\partial e_l}{\partial n^{c_i}_k}& \frac{\partial e_l}{\partial n^{c_i}_k}   \end{bmatrix} \\
     & =  \begin{bmatrix}
      - \frac{l_k(1)(l_s^T n^{c_i}_k)}{||l_k||^3}+ \frac{l_s(1)}{||l_k||} & - \frac{l_k(2)(l_s^T n^{c_i}_k)}{||l_k||^3}+ \frac{l_s(2)}{||l_k||} & \frac{1}{||l_k||} \\
      - \frac{l_k(1)(l_e^T n^{c_i}_k)}{||l_k||^3}+ \frac{l_e(1)}{||l_k||} & - \frac{l_k(2)(l_e^T n^{c_i}_k)}{||l_k||^3}+ \frac{l_e(2)}{||l_k||} & \frac{1}{||l_k||}
    \end{bmatrix}
  \end{aligned}
  \tag*{}
\end{equation}

\begin{flushright}
  (27)
\end{flushright}
\subsection{Point Line Manhattan Union Back-end Optimization}
In the back-end, we use \textit{MFs} to constrain the rotation, and structural line features to establish local and global constraints.
\subsubsection{Manhattan Constraint on Rotation}
The VIO accumulates errors as the trajectory progresses.
The \textit{MFs} provides static information that can be used to correct the rotation drift.
We define $q^M_i$ and $q^{VIO}_i$ as  the quaternions of $R^M_{b_i}$ and $R^{VIO}_{b_i}$ in frame $F_i$.
The residual of rotation is defined as follows:
\begin{equation}\label{eq16}
  e^{M}_i= 2(q^M_i \otimes (q^{VIO}_i)^{-1})_{xyz}
\end{equation}
And the Jacobian matrix $J_M$ can be obtained as:
\begin{equation}
  \begin{aligned}
    \frac{ e^{M}_i}{\partial x_i}
     & =  \begin{bmatrix}\frac{\partial e^{M}_i}{\partial q_i} & \frac{\partial e^{M}_i}{\partial t_i}\end{bmatrix} \\
     & =  \begin{bmatrix}
      [q^M_i\otimes (q^{VIO}_i)^{-1}]_{L_{xyz}} & 0
    \end{bmatrix}
  \end{aligned}
\end{equation}
\subsubsection{Structural Lines Residual}
Structural lines provide more reliable information than common lines.
Through the \textit{MFs}, we can establish both local constraints using  vanishing points in an image and global constraints using  structural lines within \textit{MW}.
Suppose  there are $n$ triangulated line features in the sliding window, with $n1$, $n2$, $n3$ lines in three respective directions.
In frame $F_i$, the \textit{MF} is expressed as $[vp_{1_i}, vp_{2_i}, vp_{3_i}]$.
We can reproject all structural lines to the image plane, and they should conform to the structural regularity in frame $F_i$.
\begin{figure}[!t]
  \centering
  \includegraphics[width=0.34\textwidth]{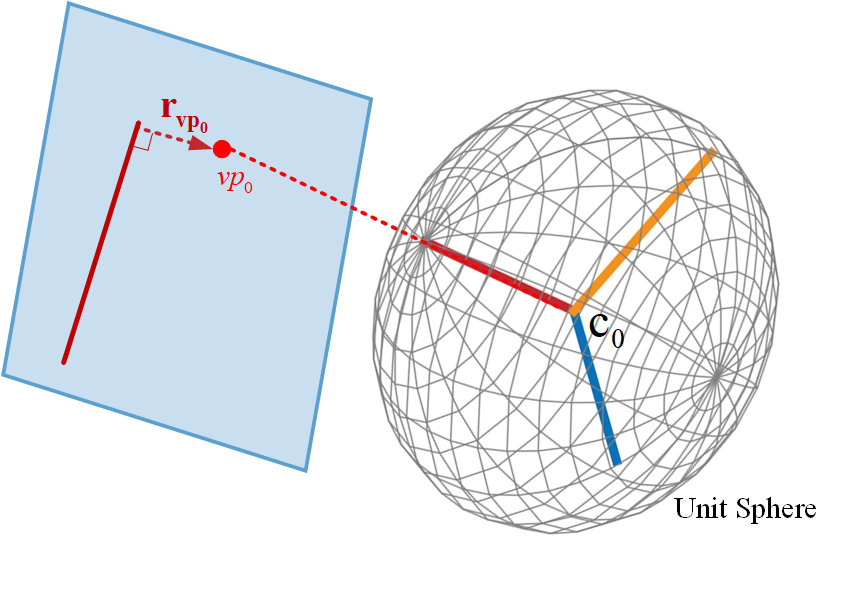}
  \caption{Representation of Manhattan residual.
    The \textit{MF} in the $c_0$ frame is shown on a unit sphere.
    The red point represents one of the vanishing points, and the red line on the image corresponds to a structural line.
    The distance between the vanishing point and the line represents the residual of the \textit{MF} in that direction.
  }
  \label{fig4}
\end{figure}
The residual in $F_i$ is given by Eq.\eqref{eq17}
\begin{equation}\label{eq17}
  e^{struct}_i=\sum_{i=1}^{n_1} d(vp_{1_i}, l) +\sum_{j=1}^{n_2} d(vp_{2_i}, l)+ \sum_{k=1}^{n_3} d(vp_{3_i}, l)
\end{equation}

Eatch frame in the sliding window can be estimated using the  equation above.
Meanwhile,  global optimization can can be performed by constraining all structural lines with the \textit{MW}.
The corressponding Jacobian $J_{struct}$ with respect to the structural lines can be obtained as follows:
\begin{equation}
  J_{struct} = \frac{\partial e^{struct}_i}{\partial n^{c_i}_k}\frac{\partial n^{c_i}_k}{\partial  L^{c_i}_k}
  \begin{bmatrix}
    \frac{\partial L^{c_i}_k}{\partial \delta x_i} & \frac{\partial L^{c_i}_k}{\partial \mathcal{L}_k} \frac{\partial \mathcal{L}_k}{\partial \delta o_i}
  \end{bmatrix}
\end{equation}
where
\begin{equation}
  \begin{aligned}
    \frac{\partial e^{struct}}{\partial n^{c_i}_k}
     & =  \begin{bmatrix}\frac{\partial e^{struct}_i}{\partial n^{c_i}_k} & \frac{\partial e^{struct}_i}{\partial n^{c_i}_k}& \frac{\partial e^{struct}_i}{\partial n^{c_i}_k}   \end{bmatrix}^T \\
     & =  \begin{bmatrix}
      - \frac{vp_i(1)(l_s^T n^{c_i}_k)}{||l_k||^3}+ \frac{vp_i(1)}{||l_k||} \\ - \frac{vp_i(2)(l_s^T n^{c_i}_k)}{||l_k||^3}+ \frac{vp_i(2)}{||l_k||} \\ \frac{1}{||l_k||} \\
    \end{bmatrix}
  \end{aligned}
\end{equation}

When there are no structural line features in the window, the back-end will automatically degenerate into a standard points-and-lines back-end.
\section{EXPERIMENTAL RESULTS}
In this section, we first evaluate our method on  benchmark datasets.
Then, real word self-recorded datasets are used to verify the system's effectiveness.
We compare our method MLINE-VINS with advanced monocular methods, such as point-based method VINS-MONO\cite{8421746}, line-based method EPLF-VINS\cite{9998999}, PL-VINS\cite{fu2020pl}, and structural constraints-based method Struct-VIO\cite{zou2019structvio}, UV-SLAM\cite{lim2022uv}.
We use the released versions and default parameters  for all testing.
All experiments are conducted on a PC with an Intel Core  i7-7700  3.60GHz CPU and 24GB of RAM.
For fair comparison, all systems are turned off loop closure and global bundle adjustment.
\subsection{Benchmark Tests}
\subsubsection{EuRoC Dataset Tests}
We first evaluate our method on the EuRoC dataset\cite{burri2016euroc}, a widely used benchmark for VIO.
This dataset is collected by a micro aerial vehicle (MAV)  equipped with stereo cameras and an IMU.
Five sequences from the machine hall meet the assumptions of the \textit{MW}.
And the GT is provided by a laser tracker.
For our evaluation, we utilize IMU data and images from the left-camera images  to test our proposed method.

To evalute the performance of eatch method, we use the root meam squared error (RMSE) as the evaluation metric.
For eatch sequence, the average RMSE is calculated over 10 runs.
\begin{table}[!h]
  \centering
  \caption{TRANSLATION RMSE ATE WITHOUST LOOP CLOUSURE of EuRoc DATASET(UNIT: M)}
  \label{tab1}
  \begin{tabular}{cccccc}
    \toprule
    \specialrule{0em}{3pt}{3pt}
    Method                            & MH01             & MH02             & MH03             & MH04             & MH05             \\
    \specialrule{0em}{3pt}{3pt}
    \hline
    \specialrule{0em}{1pt}{1pt}
    VINS-MONO\cite{8421746}           & 0.202            & 0.188            & 0.228            & 0.370            & 0.297            \\
    \specialrule{0em}{1pt}{1pt}
    Struct-VIO\cite{zou2019structvio} & {0.079}          & 0.145            & {0.103}          & {\textbf{0.130}} & 0.182            \\
    \specialrule{0em}{1pt}{1pt}
    PL-VINS\cite{fu2020pl}            & 0.216            & 0.193            & 0.218            & 0.247            & 0.325            \\
    \specialrule{0em}{1pt}{1pt}
    EPLF-VINS\cite{9998999}           & 0.140            & {0.088 }         & 0.114            & {0.182}          & {0.182}          \\
    \specialrule{0em}{1pt}{1pt}
    UV-SLAM\cite{lim2022uv}           & 0.139            & 0.094            & 0.189            & 0.261            & 0.188            \\
    \specialrule{0em}{1pt}{1pt}
    MLINE-VINS                        & {\textbf{0.074}} & {\textbf{0.072}} & {\textbf{0.093}} & 0.196            & {\textbf{0.130}} \\
    \specialrule{0em}{1pt}{1pt}
    \bottomrule
  \end{tabular}
\end{table}

Table \ref{tab1} presents the absolute translation error (ATE) results of our method compared to other approaches, with results sourced from their original papers.
First, methods incorporating line features, such as PL-VINS, EPLF-VINS and MLINE-VINS, generally demonstrate higher accuracy than VINS-MONO.
Second, EPLF-VINS and MLINE-VINS which utilize line tracking methods, tend to outperform PL-VINS.
Additionally, methods incorporating structural constraints generally achieve better performance than conventional point-and-line-based approaches.
By leveraging novel structural constraints in local and global optimization, our method demonstrates higher accuracy compared to UV-SLAM and Struct-VIO.
Overall, the proposed method achieves the best performance, across most sequences, except for \textit{MH04}.
However, the performance of our system heavily depends  on proper initialization.
In particular, inaccuracies in the initial gyro bias  occasionally degrade system performance.
For the \textit{MH02}, our method shows a  $50.3\%$ improvement in accuracy over Struct-VIO.
These results highlight the effectiveness of our approach.

\subsubsection{KAIST-VIO Dataset Tests}
To evalute the robustness of our system, we test our method on the KAIST-VIO dataset\cite{jeon2021run}.
The dataset is collected by a UAV platform equipped with a monocular RGB camera(30 Hz) and IMU(100 Hz).
It includes four   different sports, each with three modes: normal, fast and rotation, which are challenging for VIO systems.
We compare our method with VIN-MONO\cite{8421746}, EPLF-VINS\cite{9998999}, PL-VINS\cite{fu2020pl} and UV-SLAM\cite{lim2022uv}.
All other operations are the same as those used in the EuRoc dataset tests.

Table \ref{tab2} shows the ATE RMSE results of all methods.
Compared to other methods, our method demonstrates superior robustness.
In particular, other methods tend to fail during rapid rotation  scenarios due to difficulties in estimating relationships between frames.
Our method operates reliably thanks to the local and global Manhattan constraints.
It is important to  note that point and line features  can sometimes be unreliable in a rapidly changing scenarios, as triangulation may fail during quick camera rotations.
With the aid of \textit{MFs}, the UV-SLAM and MLINE-VINS can  incorporate additional constraints in optimization.
Besides, by leveraging the novel line tracking method and Manhattan constraints optimization, our method achieves heigher accuracy.

\begin{table*}[!ht]
  \centering
  \caption{TRANSLATION RMSE ATE WITHOUT LOOP CLOSURE of KAIST-VIO DATASET (UNIT: M)}
  \label{tab2}
  \begin{tabular}{cccccccccccc}
    \toprule
    \specialrule{0em}{3pt}{3pt}
    Method                  & cir\_f           & cir\_n         & cir\_r           & inf\_f           & inf\_n           & inf\_r           & rot\_f           & rot\_n           & squ\_f           & squ\_n         & squ\_r           \\
    \specialrule{0em}{3pt}{3pt}
    \hline
    \specialrule{0em}{1pt}{1pt}
    VINS-MONO\cite{8421746} & 0.126            & 0.234          & {0.257}          & {0.080}          & 0.055            & {0.360}          & 0.337            & ---              & {\textbf{0.061}} & {0.077}        & ---              \\
    \specialrule{0em}{1pt}{1pt}
    PL-VINS\cite{fu2020pl}  & 0.119            & 0.122          & 0.305            & {\textbf{0.063}} & 0.057            & 0.415            & 0.527            & ---              & {0.062}          & 0.145          & {0.232}          \\
    \specialrule{0em}{1pt}{1pt}
    EPLF-VINS\cite{9998999} & 0.144            & {0.069}        & 0.437            & 0.234            & 0.129            & 0.461            & {0.206}          & ---              & 0.094            & 0.097          & 0.308            \\
    \specialrule{0em}{1pt}{1pt}
    UV-SLAM\cite{lim2022uv} & {0.100}          & {0.100}        & 0.324            & 0.086            & {0.054}          & 0.382            & 0.473            & ---              & 0.077            & {0.079}        & ---              \\
    \specialrule{0em}{1pt}{1pt}
    MLINE-VINS(ours)        & {\textbf{0.095}} & \textbf{0.051} & {\textbf{0.118}} & 0.130            & {\textbf{0.051}} & {\textbf{0.306}} & {\textbf{0.143}} & {\textbf{0.212}} & 0.087            & \textbf{0.066} & {\textbf{0.206}} \\
    \specialrule{0em}{1pt}{1pt}
    \bottomrule
  \end{tabular}
\end{table*}

\subsection{Qualitative Evaluation}
To better evaluate our method, we conduct several  qualitative evaluation experiments.
\subsubsection{Evaluation of Line Segment Tracking}
\begin{figure}[!ht]
  \centering
  \includegraphics[width=3.5in]{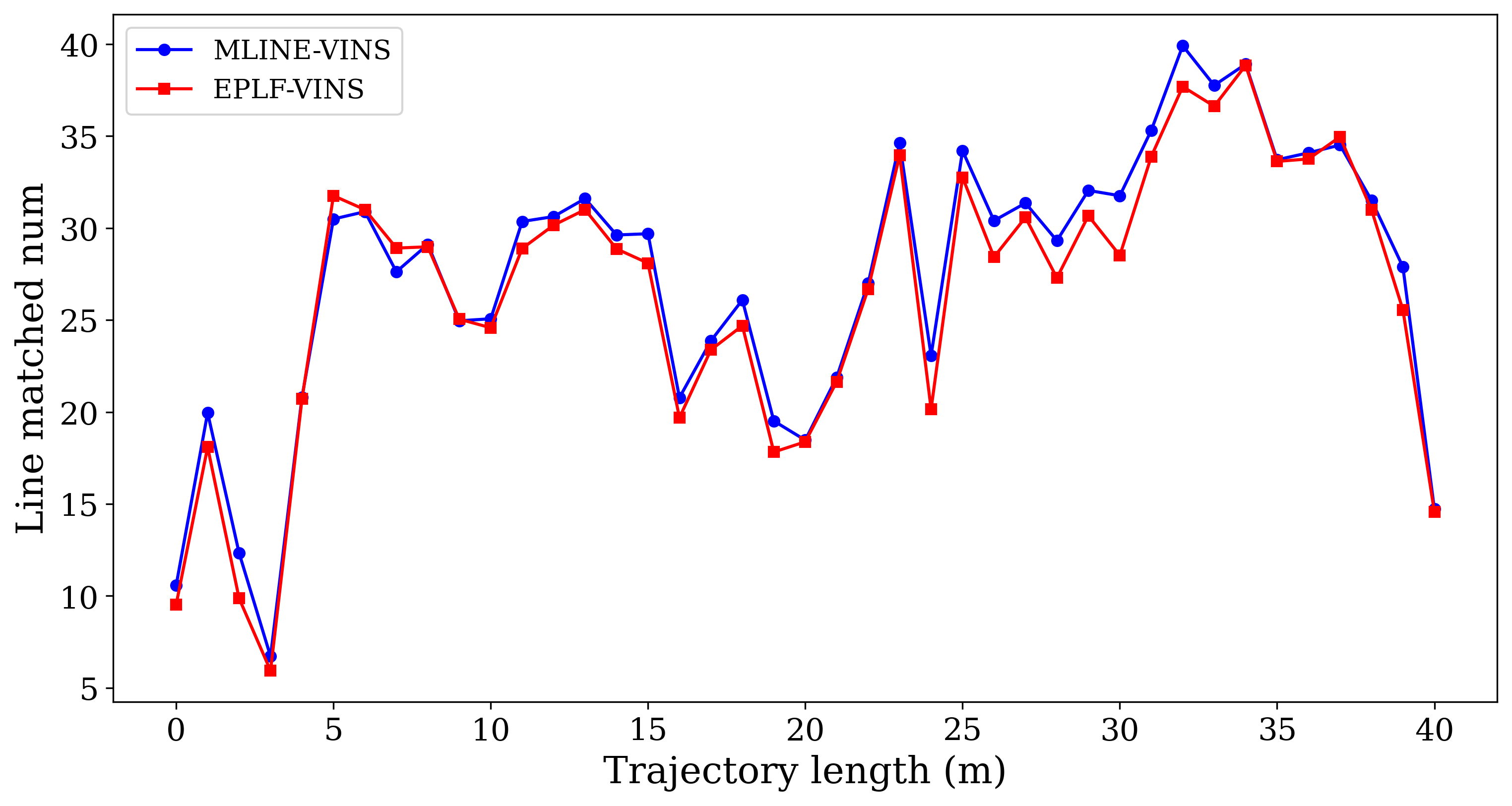}
  \caption{Average number of line matches  between MLINE-VINS and EPLF-VINS in EuRoC tests.
    The blue line represents the number of lines tracked by MLINE-VINS (ours), and the red line represents the number of lines tracked by EPLF-VINS.}
  \label{fig9}
\end{figure}

We evaluate the number of line features tracked across consecutive frames by comparing  our method with EPLF-VINS on the EuRoc dataset, as shown in Fig.\ref{fig9}.
By accounting for changes in line length during modeling, our method identifies more matching lines than EPLF-VINS.

\subsubsection{Evaluation of Drift in Challenging Environment}
\begin{figure}[!h]
  \centering
  \includegraphics[width=0.48\textwidth]{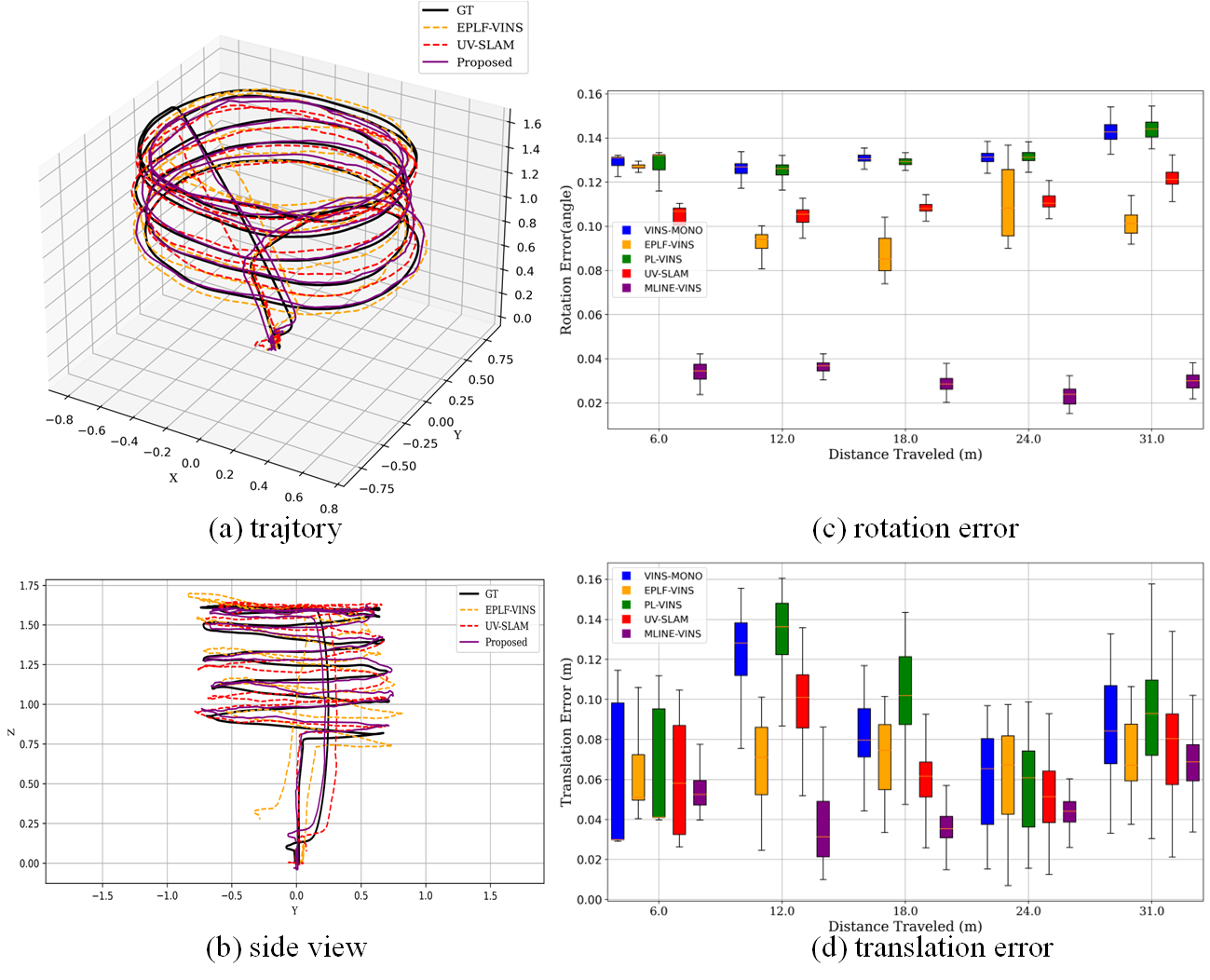}
  \caption{Trajectory and relative pose error on the KAIST-VIO test.
    (a) and (b) show the trajectory of our method, EPLF-VINS and UV-SLAM.
    (c) and (d) show the average relative pose error for rotation and translation of each method.
  }
  \label{trajtory_ape}
\end{figure}

To assess the drift of the proposed method, we compare the cumulative error across different methods using the  KAIST-VIO dataset.
Fig.\ref{trajtory_ape} presents  the tracking trajectory and  drift errors for each method on the \textit{cir\_n} sequence.
As shown, incorporating additional line features significantly reduces drift compared to only point-based methods, particularly during rotation.
Compared to simple point and line features, structural regularity offers a more reliable means  of constraining drift in both translation and rotation.
By leveraging novel line feature tracking and  Manhattan constraints, our method achieves the best performance with the least drift.

\subsubsection{Evaluation of Vanishing Point Estimation}
\begin{figure}[!t]
  \centering
  \includegraphics[width=0.48\textwidth]{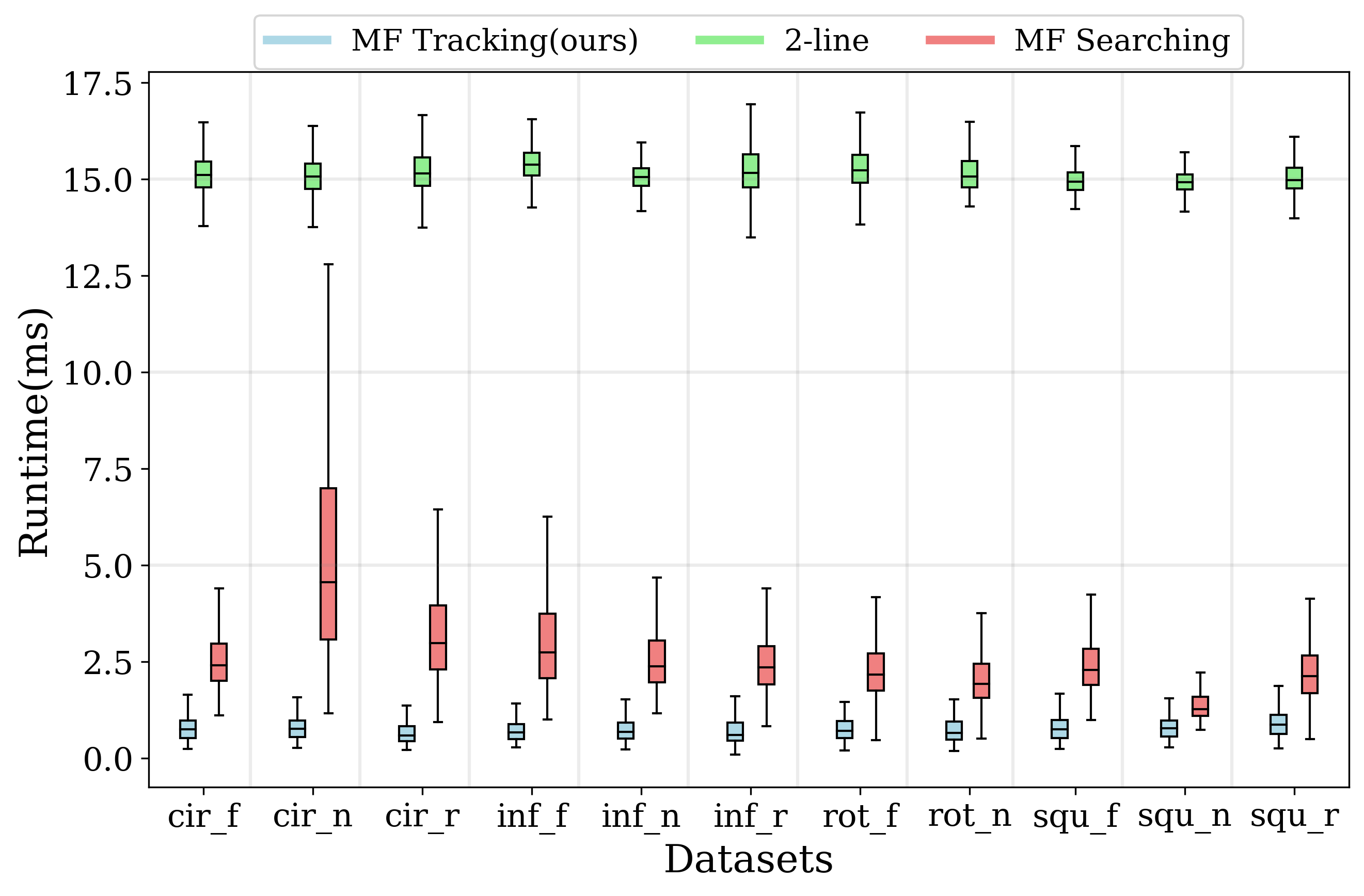}
  \caption{Runtimes of \textit{MF} estimation using MF Tracking (ours), 2-line and MF Searching in KAIST-VIO dataset}
  \label{MW_runtime}
\end{figure}

We compare the runtime  of different Manhattan estimation methods, including VPs estimatinon and  structural lines  classification, as shown in Fig.\ref{MW_runtime}.
The 2-line method\cite{7926628}  requires the most time because it repeatedly executes RANSAC across all  three directions.
While MF Searching\cite{li2020quasi} is faster than the 2-line method, its performance  is more sensitive to the number and distribution of lines.
Benefiting from  the proposed tracking-by-detection module for \textit{Mfs} estimation, the proposed method achieves superior efficiency compared to the advanced approaches.

\begin{figure}[!t]
  \centering
  \includegraphics[width=0.48\textwidth]{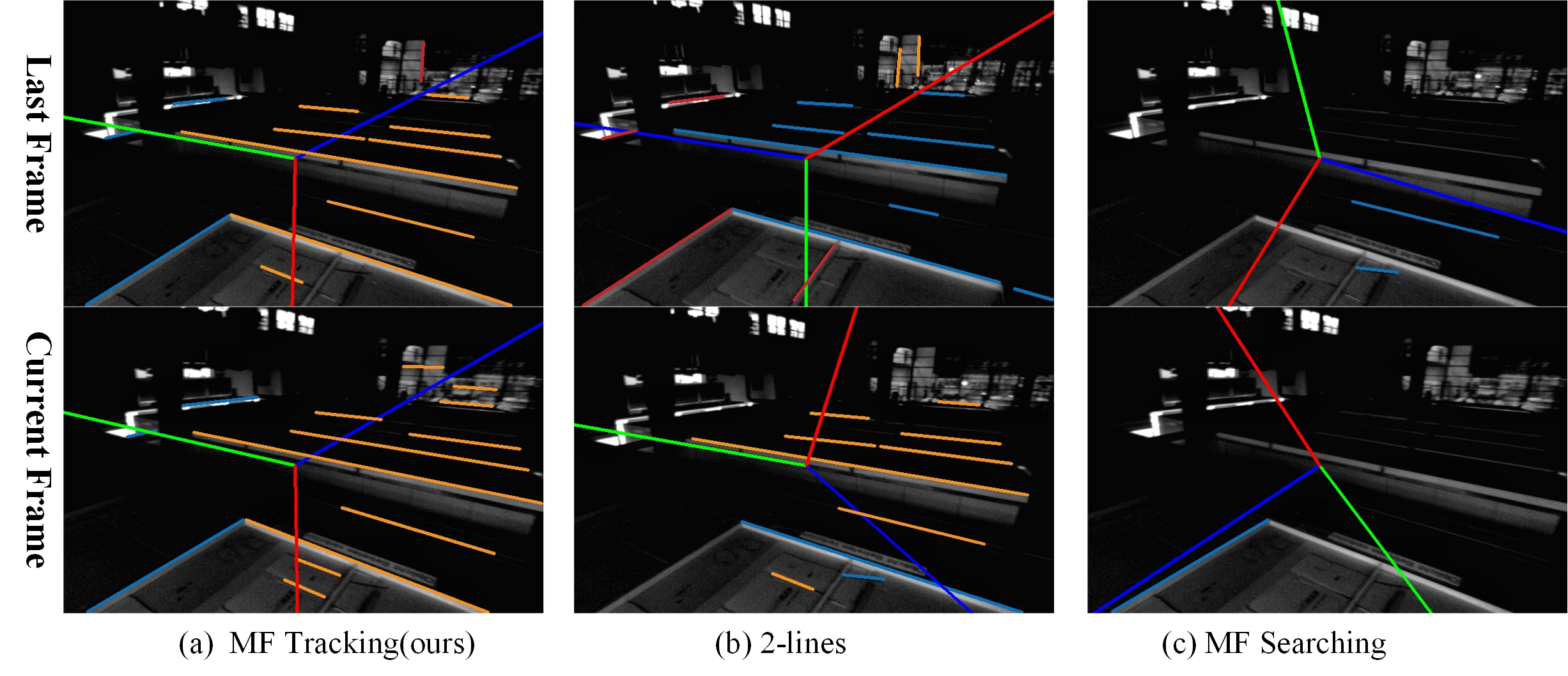}
  \caption{\textit{MFs} estimation in consecutive frames.
    (a), (b) and (c) show the results of \textit{MFs} estimatinon  using  the Tracking-by-Detection Strategy (ours), the 2-line method and the MF Searching.
    The top image  shows the last frame, and the bottom image shows the current frame.
  }
  \label{Manhattan_performance}
\end{figure}

We also evaluate the performance of different \textit{MF} estimation methods  per frame, as shown in Fig.\ref{Manhattan_performance}.
The MF Searching struggles to provide  accurate  \textit{MF}  estimations when line features are  insufficient.
The 2-line method fails under uneven line distribution.
Besides, both the MF Searching and the 2-line method encounter issues with random and disordered Manhattan axes.
In contrast, our method demonstrates stable and accurate performance across consecutive frames, making it suitable for real-world applications.

\subsection{Real World Tests}
\begin{figure*}[!ht]
  \centering

  \includegraphics[width=0.99\textwidth]{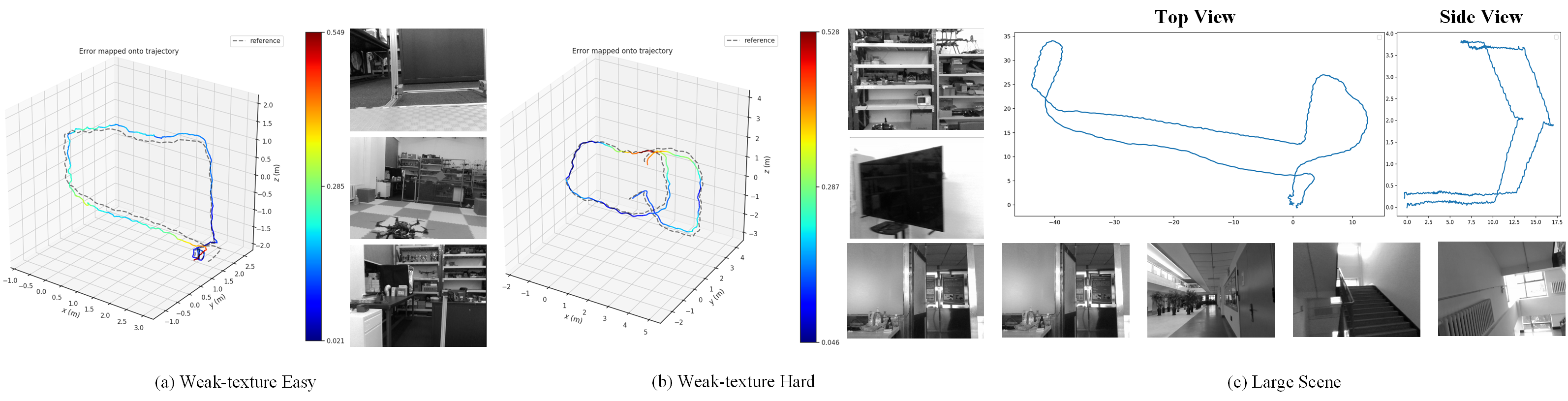}
  \caption{Our collected data for evaluation  VIO methods.
    For each sequence, we present the corresponding trajectory visualizations (left) and sample images (right).
    The datasets include (a) \textit{Weak-texture Easy }and (b) \textit{Weak-texture Hard} sequences captured in a small indoor environment equipped with motion capture systems for ground truth measurements and (c)  \textit{Large Scene} sequence containing long corridor scenes.
    These datasets cover various challenging scenarios, including motion blur, texture-deficient surfaces, and pure rotational movements.}
  \label{collected_dataset}
\end{figure*}

In this section, we present real-world experiments to evaluate the performance of the proposed method in various indoor scenes.
We use D435I camera and its internal IMU to collect RGB images and IMU messages for evaluation.
RGB images are 640$\times$480 recorded at 30 Hz and IMU messages are at 200 Hz.
The camera and IMU are calibrated using Kalibr\cite{furgale2013unified} and Imu\_Utils\cite{woodman2007introduction}.
In these experiments, all features including points and lines, are extracted directly from the original images.
The datasets used are illustrated in Fig.\ref{collected_dataset}.

\subsubsection{Small Indoor Scene}
First, the proposed method is evaluated in a small indoor laboratory setting.
As shown in Fig.\ref{collected_dataset}, the data is collected under two distinct conditions: slow rotation (referred to as \textit{Weak-texture Easy}) and rapid rotation (referred to as \textit{Weak-texture Hard}).
We use motion capture equipment to obtain  ground truth data and RMSE to evaluate  performance.

For other methods, the default parameters are used in the experiments.
The back-end of UV-SLAM does not perform well in real-world scenarios, resulting in  failure across all sequences.
The RMSE results are shown in Table \ref{tab3}.
MLINE-VINS achieves the best performance on both \textit{Weak-texture Easy} and \textit{Weak-texture Hard} datasets.
On the \textit{Weak-texture Hard} dataset, which includes pure rotation and fast movements, traditional line-matching methods occasionally fail.


\begin{table}[!ht]
  \centering
  \caption{TRANSLATION RMSE ATE WITHOUST LOOP  CLOSURE of COLLECTED DATASET(UNIT: M)}
  \label{tab3}
  \begin{tabular}{ccccc}
    \toprule
    \specialrule{0em}{3pt}{3pt}
    Method            & \makecell{VINS-                                  \\MONO} &  \makecell{PL- \\VINS} &  \makecell{EPLF- \\VINS}         & \makecell{MLINE- \\VINS(ours)} \\
    \specialrule{0em}{3pt}{3pt}
    \hline
    \specialrule{0em}{1pt}{1pt}
    Weak-texture Easy & 0.316           & 0.218 & 0.357 & \textbf{0.216} \\
    \specialrule{0em}{1pt}{1pt}
    Weak-texture Hard & 0.365           & 0.528 & 0.248 & \textbf{0.227} \\
    \specialrule{0em}{1pt}{1pt}
    Average           & 0.341           & 0.373 & 0.302 & \textbf{0.221} \\
    \bottomrule
  \end{tabular}
\end{table}


\subsubsection{Large Indoor Scene}
To further evaluate our method in real-world scenarios, we conducted experiments in a large indoor environment.
In this setting, the camera traverses the first and second floors,  eventually returning to the starting point.
During this processs, there are many challenging situations, including motion blur and weakly textured scenes as illustrated in Fig.\ref{collected_dataset}(c).

\begin{figure}[!h]
  \centering
  \includegraphics[width=0.45\textwidth]{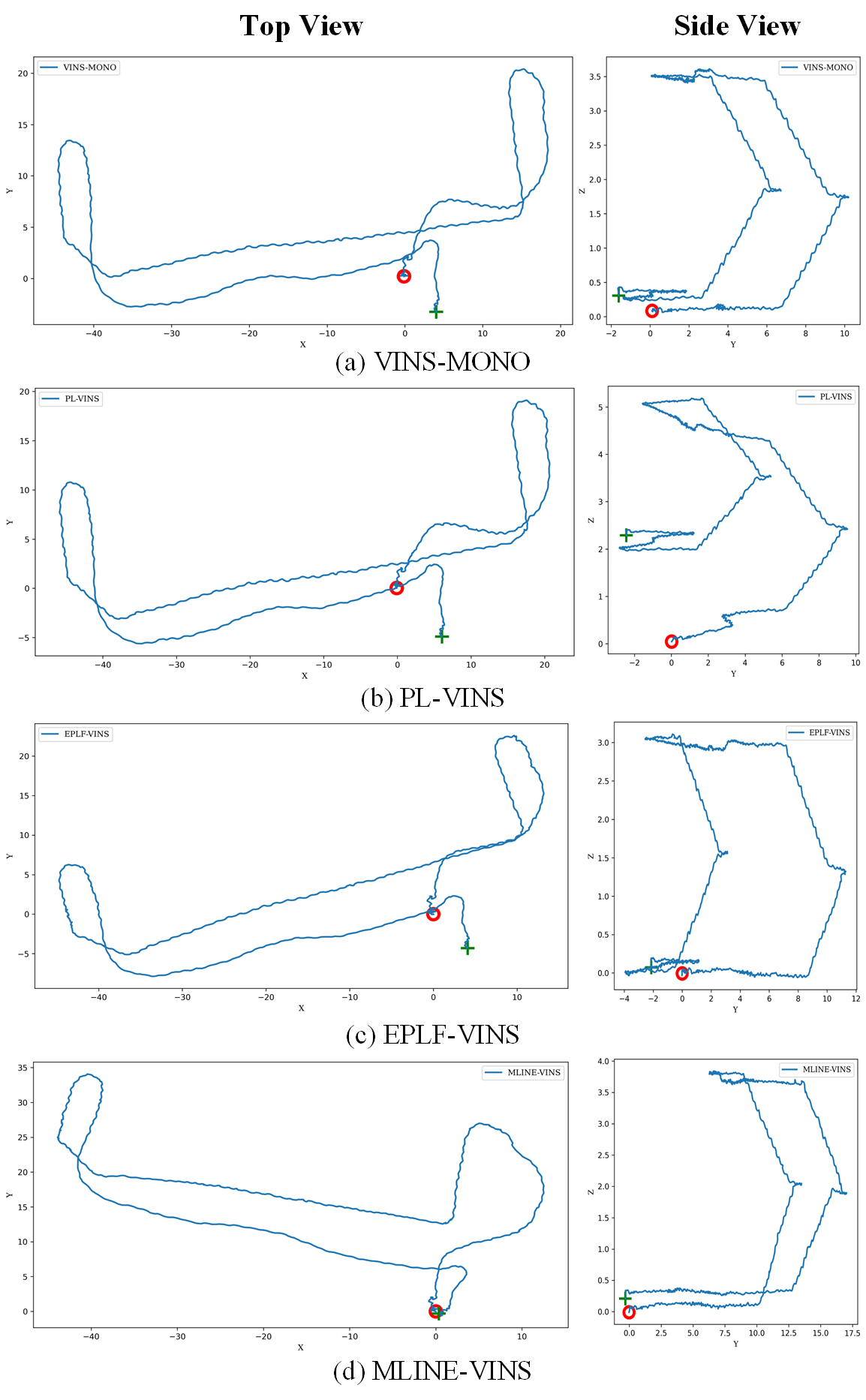}
  \caption{Results of \textit{Large Scene} test.
    The left image shows the top view of the trajectory, while the right image displays the side view..
    The red cycle represents the starting point, and the green cross marks  the endpoint.
  }
  \label{large_trajectory}
\end{figure}

Unlike the small indoor scene, we evaluate the performance of each method based on the drift between the start and end points, which are at the same position
Since  UV-SLAM cannot complete the entire trajectory, its results omitted.
The results, shown in Fig.\ref{large_trajectory}, indicate that MLINE-VINS demonstrates superior performance in suppressing long-term accumulated errors.
By incorporating Manhattan constraints  into both local and global optimization, the proposed method achieves the best performance with minimal drift error.


\subsection{Runtime Analysis}
The computational performance of each module was evaluated on a per-frame basis, as shown in  Table \ref{tab5}.
Our method and EPLF-VINS demonstrated superior efficiency in line  feature  detection and matching, primarily due to the implemented line tracking algorithm.
Furthermore, the proposed tracking-by-detection module for \textit{MFs} estimation proved to be more computationally efficient than the  2-lines method employed in UV-SLAM.
Despite incorporating additional line features and Manhattan constraints, the Local VIO is not burdened significantly.
Although the runtime of our method is slightly higher than that of  VINS-MONO, it is still acceptable for real-time applications.
\begin{table}[!htbp]
  \centering
  \caption{RUNTIME of EACH MODULE(ms)}
  \label{tab5}
  \resizebox{\textwidth/2}{!}{ 
    \begin{tabular}{cccccccc}
      \toprule
      \specialrule{0em}{3pt}{3pt}
      Thread & Module                        & \makecell{VINS-                                 \\MONO}      & \makecell{PL-\\VINS}        & \makecell{EPLF-\\VINS} & \makecell{UV-\\SLAM} & \makecell{MLINE-\\VINS} \\
      \specialrule{0em}{3pt}{3pt}
      \hline
      \specialrule{0em}{1pt}{1pt}
      1      & \makecell{Points  Detection                                                     \\\&Matching}    &  13.15         &       13.71 &13.34&13.56 &13.29                                     \\
      \specialrule{0em}{1pt}{1pt}
      2      & \makecell{Lines Detection                                                       \\\& Matching}    &---           & 33.03  &14.37&22.09     &14.80                                     \\
      \specialrule{0em}{1pt}{1pt}
      3      & \makecell{Manhattan Detection                                                   \\\& Matching} & ---          & ---     &---&16.47    &4.73                                  \\
      \specialrule{0em}{1pt}{1pt}
      4      & Local VIO                     & 41.12           & 52.50 & 51.55 & 86.61 & 61.13 \\
      \specialrule{0em}{1pt}{1pt}

      \specialrule{0em}{1pt}{1pt}
      \specialrule{0em}{1pt}{1pt}
      \bottomrule
    \end{tabular}
  }
\end{table}

\section{Conclusion}
In this paper, we introduce MLINE-VINS, a novel visual-inertial odometry system that integrates line features and Manhattan World  constraints.
To achieve real-time performance, we introduce  a novel line feature tracking algorithm and tracking-by-detection module, which efficiently tracks variable-length lines and detects Manhattan frame across consecutive images.
To simplify coordinate transformations, we align the VIO world frame with the Manhattan World framework.
A pose-guided Manhattan frame verification mechanism is employed to ensure the reliability of the Manhattan frames.
By incorporating structural lines and Manhattan frame constraints, we propose a novel back-end optimization framework that incorporates both local and global constraints to enhance the system' s accuracy and robustness.
Comprehensive evaluations on benchmark and custom datasets demonstrate that MLINE-VINS achieves advanced performance in terms of both accuracy and robustness.
In future work, we will focus on leveraging line features to further enhance the system's robustness and accuracy, particularly in challenging outdoor environments



\bibliographystyle{IEEEtran}
\bibliography{reference}

\end{document}